\documentclass[11pt]{article}
\usepackage[a4paper,margin=2.3cm]{geometry}
\usepackage[T1]{fontenc}
\usepackage{lmodern,microtype,graphicx,amsmath,amssymb,dsfont,booktabs,xcolor,pifont}
\usepackage[labelfont=bf,labelsep=space,font=small,justification=justified,singlelinecheck=false]{caption}
\usepackage[super,comma,sort&compress]{natbib}

\usepackage[hidelinks]{hyperref}
\usepackage{lineno}
\usepackage{flafter}
\graphicspath{{figures/}}
\DeclareCaptionLabelSeparator{bar}{ $|$ }
\newcommand{\figpage}[2]{\makebox[\textwidth][c]{\includegraphics[width=#1]{#2}}}
\newcommand{\mainfig}[1]{\centering\includegraphics[width=\textwidth,height=0.6\textheight,keepaspectratio]{#1}}

\newcommand{\Xhi}{X$_{\mathrm{hi}}$}
\newcommand{\Xlo}{X$_{\mathrm{lo}}$}

\newcommand{\capTitleA}{Rhythm is common to real and rewired wiring; antagonist coordination is strongest in the real wiring.}
\newcommand{\capBodyA}{\textbf{a}, Question and system (schematic): leg-motor subnetworks of two male fly ventral-nerve-cord connectomes (MaleCNS, 10,173 neurons, 462,650 connections; MANC, 10,440 and 634,207); dots, the four joints of a leg. \textbf{b}, Approach (schematic): a rate network with the signed synapse counts as fixed weights, driven through the DNg100 descending neurons; 20 extensor--flexor pool pairs are scored for rhythmicity and for antagonist coordination S (mean $-$corr(extensor, flexor)). Thirty rewired networks per connectome (six families of five, preserving progressively more of the real structure) receive the same protocol and pre-specified claim levels (Supplementary Fig. 1). \textbf{c}, Rhythmicity (left) and S (right) of the real networks (bars; dark teal, MaleCNS; teal, MANC) and the rewired networks (points, coloured by family), each at its own frozen setting (60-s confirmation, 15--25 s, mean of 20 noisy runs). 9 and 14 of 30 rewired networks are at least as rhythmic as the real networks, but none reaches their S (0.31 against at most 0.14; 0.17 against at most 0.08; levels strong and moderate). \textbf{d}, Reciprocal innervation (schematic): for example, a premotor neuron excites one pool and, through an inhibitory neuron, inhibits its antagonist, so the pools alternate. Bottom, reciprocal-innervation index (real 0.48 and 0.38; all rewired networks below 0). Right, S relative to the real network after premotor input is moved across antagonistic pools with motor neurons' input strength nearly unchanged (median change below 4\%): within-pool controls (x = 0), strength-matched and dose-matched cross-pool reassignments (points, networks; lines, medians; 15--25 s; Fig. 4). \textbf{e}, Joint scores of the real networks and of the best rewired network (lines), 15--25 s; only the thorax--coxa joint is `strong' in both connectomes in all four windows (Supplementary Fig. 4). Source data are provided as a Source Data file.}
\newcommand{\capTitleB}{Rewired networks produce rhythm but weaker antagonist coordination.}
\newcommand{\capBodyB}{\textbf{a}, Example activity of the six thorax--coxa pairs in the real MaleCNS network and in the rewired network with the highest confirmation score (lineage block), each at its own frozen setting (example simulation with a CPU re-implementation of the model; not a confirmation run; S = 0.32 and 0.16). Each pair is scaled by its larger pool s.d.; r, extensor--flexor correlation over 15--25 s; n.m., amplitude gate not met. \textbf{b,c}, Parameter scan: best S against maximum rhythmicity over the eligible settings of the 132-setting scan (one 25-s simulation per setting). Shaded, rhythmicity above the real network's maximum. \textbf{d--g}, 60-s confirmation at each network's frozen setting: rewired networks (points, mean $\pm$ s.e.m. of 20 runs; hollow, secondary MaleCNS families that also preserve reciprocal connections; diamond, a network with 9 of 20 undefined runs, mean of 11), family medians (dark ticks), real network (line; band $\pm$ s.e.m.) and the margin required of every family median for the pre-specified `strong' level (dotted). Panel titles give the level reached. \textbf{h,i}, S of every confirmation run of the real networks; amber, runs that entered a fast state ($\geq$ 3 Hz with $\leq$ 3 modulated pairs; open circles) after 15--25 s; dashed, noise-free runs; thick line, mean of 20 noisy runs. The 300-s continuation is shown in Supplementary Fig. 2. Source data are provided as a Source Data file.}
\newcommand{\capTitleC}{Reciprocal innervation of antagonistic pools is specific to the real wiring.}
\newcommand{\capBodyC}{\textbf{a}, The reciprocal-innervation index on data: each point is one neuron's signed one- plus two-step influence on the extensor pool (x) and on the flexor pool (y) of the MaleCNS T2L thorax--coxa pair (one of the two pairs defining the real network's median index), in the real network and in leg-block rewired network 0 (symmetric-log axes, linear within $\pm$0.01); index = $-$Pearson r across neurons (0.48 and $-$0.34). \textbf{b}, Network-level index (median over the 20 pairs) of every rewired network (points; dark ticks, family medians) and of the real networks (lines). \textbf{c}, Pair by pair: real index (circles; filled, above all 30 main rewired networks) against the range (bars) and median of the 30 rewired networks; 19 of 20 pairs in MaleCNS, 18 of 20 in MANC. \textbf{d}, Real per-pair index in the two connectomes; Spearman $\rho$ = 0.82 (n = 20 pairs). Source data are provided as a Source Data file.}
\newcommand{\capTitleD}{Reassigning premotor inputs across antagonistic pools abolishes coordination.}
\newcommand{\capBodyD}{\textbf{a}, Reassignment (schematic): edges onto motor neurons exchange targets within a pool (within-pool control) or between the extensor and flexor pools of the same leg and joint (cross-pool); strength-matched versions exchange only edges of the same sign and synapse-count decile. \textbf{b}, Example activity of the six thorax--coxa pairs in the real MaleCNS network and in cross-pool network 0 at the real frozen setting (example simulation as in Fig. 2a). \textbf{c,d}, S relative to the real network in the same window (R above each panel) for every reassigned network at the real frozen setting: original, strength-matched, and dose-matched reassignments (labels: designed net fraction of input moved across pools; MaleCNS 22\% and 27\%, MANC 20\% and 25\%). Filled circles, 15--25 s; open diamonds, 50--60 s; bars, medians of five networks; error bars, s.e.m. of 20 runs divided by R; lines at R, 0.8 R (within-pool criterion) and 0.5 R (cross-pool criterion). \textbf{e}, Cross-pool arms relative to the real network: S, rhythmicity and the fraction of oscillating motor neurons, for the original reassignment and the dose-matched one (medians of five networks; circles MaleCNS, squares MANC; filled 15--25 s, open 50--60 s); shaded, pre-specified criterion met. \textbf{f}, Extensor--flexor correlation of the pairs that contribute in the real networks (MaleCNS 9, MANC 8), real (bars) and within-pool (mint) or original cross-pool (dark red) networks (hollow, pair modulated in fewer than half the runs); 8 of 9 and 7 of 8 pairs stay modulated after cross-pool reassignment. \textbf{g}, S relative to the real network against the net fraction of premotor input moved across antagonistic pools, every network (15--25 s); dashed, medians of the strength-matched cross-pool series; circles MaleCNS, squares MANC. Source data are provided as a Source Data file.}
\newcommand{\capTitleE}{Wiring-specific coordination is concentrated at the thorax--coxa joint.}
\newcommand{\capBodyE}{\textbf{a}, Score of each antagonistic pair in the real network (fill; mean of 20 runs, 15--25 s) drawn at its joint on a schematic fly, MaleCNS and MANC; ring, the real network exceeds all 30 main rewired networks for that pair (6 of 20 and 4 of 20 pairs); $\times$, joint without an antagonistic pair. \textbf{b}, Mean pair score per joint for the real network (bar; pale band $\pm$ s.e.m.) and the 30 rewired networks (points sorted within each joint, coloured by family; $\pm$ s.e.m.). \textbf{c}, Pre-specified readouts: S over all 20 pairs, the 16 pairs of exactly assigned motor neurons, and the 14 pairs outside the thorax--coxa joint; squares, pre-specified level at 15--25 s (left) and 50--60 s (right). The thorax--coxa joint alone is `strong' in both connectomes and all four windows (Supplementary Fig. 4). Source data are provided as a Source Data file.}
\newcommand{\capTitleEDA}{Model, rewired networks and protocol.}
\newcommand{\capBodyEDA}{\textbf{a}, Model (schematic). The leg-motor subnetwork of each connectome is simulated as a rate network with fixed connection weights (equation; W, signed synapse counts; g, gain; $\beta$, inhibition scale) driven only by a sustained input d to the left and right DNg100 descending neurons. Motor neurons are grouped by annotated target muscle into extensor (coral) and flexor (green) pools at four joints of each leg: thorax--coxa (ThC), coxa--trochanter (CTr), femur--tibia (FTi) and tibia--tarsus (TiTa). The antagonist score S averages, over the 20 antagonistic pool pairs, $-$corr(extensor, flexor) for pairs in which both pools are modulated and 0 otherwise. \textbf{b}, Composition of the extracted subnetworks: MaleCNS 10,173 neurons and 462,650 connections; MANC 10,440 neurons and 634,207 connections. Lower bars, motor neurons with an exact or approximate joint assignment, or unmapped. \textbf{c}, The six families of rewired networks and the structure each preserves (filled circles); five networks per family. \textbf{d}, Protocol. Every network is simulated at the same 132 settings (11 gains $\times$ 3 inhibition scales $\times$ 4 drives; 25-s runs scored at 15--25 s); shown are the scores of the real MaleCNS network at its 27 eligible settings (data; sand, not eligible; amber ring, selected setting). The best eligible setting is frozen and re-simulated for 60 s with 20 new noise realizations and 3 noise-free initial states, scored in four windows. Source data are provided as a Source Data file.}
\newcommand{\capTitleEDB}{The coordinated state over five minutes.}
\newcommand{\capBodyEDB}{300-s simulations at the frozen settings of the real MANC network, its two strongest rewired networks (lineage block 1 and 0) and the real MaleCNS network; 40 noisy trajectories per network (20 continuing the confirmation runs, 20 new) and 10 noise-free trajectories. A 10-s window counts as coordinated when S $\geq$ 0.10 with at least six modulated pairs (pre-registered). \textbf{a}, Fraction of noisy trajectories in the coordinated state. \textbf{b}, Mean S (undefined S counted as 0; band, $\pm$ s.e.m.). Sand band, the last window (290--300 s). \textbf{c}, State of every trajectory in every window (filled, coordinated; below the line, noise-free trajectories). \textbf{d}, Switching rate: departures from the coordinated state (followed by at least three non-coordinated windows) per minute spent coordinated, with exact Poisson 95\% confidence intervals; MANC real 0.24 per minute (mean dwell 4.2 min), lineage block 1 3.7 and lineage block 0 5.6; MaleCNS real, no departure. \textbf{e}, S of every noisy trajectory at 290--300 s; bars, means. Source data are provided as a Source Data file.}
\newcommand{\capTitleEDC}{Dose-matched reassignment of premotor inputs.}
\newcommand{\capBodyEDC}{\textbf{a}, Design check: net fraction of premotor input moved across antagonistic pools against the change in the motor neurons' input strength (median relative change, larger of excitatory and inhibitory), every reassigned network; hollow, strength-matched constructions (exchanges restricted to the same sign and synapse-count decile); W, dose-matched within-pool control; X$_{\mathrm{lo}}$ and X$_{\mathrm{hi}}$, dose-matched cross-pool arms. Right, each dose-matched network's median change (marker) and 95th percentile (bar top), larger of excitatory and inhibitory; across both connectomes and both input signs the medians are 1.0--2.3\% (W), 1.5--2.6\% (X$_{\mathrm{lo}}$) and 2.0--3.9\% (X$_{\mathrm{hi}}$) and the 95th percentiles 9--19\%, 9--20\% and 11--30\%, so the within-pool control, which kept coordination, changed its tail by a similar amount to X$_{\mathrm{lo}}$, which lost most of it. \textbf{b}, Pre-registered conditions for a selective loss (filled, met): primary test X$_{\mathrm{hi}}$ against W at the real network's frozen setting; secondary, X$_{\mathrm{hi}}$ at re-optimised settings and X$_{\mathrm{lo}}$ at the real setting. \textbf{c}, S of W and X$_{\mathrm{hi}}$ at their own re-optimised frozen settings (filled, 15--25 s; open, 50--60 s; mean $\pm$ s.e.m. of 20 runs; lines, real network). \textbf{d}, Activity maintenance relative to the real network: fraction of oscillating motor neurons (circles) and of the real network's contributing pairs still modulated (triangles), medians of five networks; shaded, criterion ($\geq$ 0.8) met. Source data are provided as a Source Data file.}
\newcommand{\capTitleEDD}{Where and when the advantage of the real wiring holds.}
\newcommand{\capBodyEDD}{\textbf{a}, Pre-specified level (dark teal strong, light teal moderate, white weak) of each readout in each confirmation window, with the real network's value: S over all 20 pairs, over the 16 pairs of exactly assigned motor neurons, over the 14 pairs outside the thorax--coxa joint, and each joint alone. \textbf{b}, Score of every antagonistic pair of the real networks in the four windows (dot, modulated in at least half of the runs); in MANC the contributing pairs shift over time towards the thorax--coxa joints. Source data are provided as a Source Data file.}

\title{\textbf{How much of fly walking is written in the wiring?}}
\author{Isabel Guan$^{1,*}$, Yuntian Zhao$^{2}$, Dingyuan Zhang$^{3}$, Shipeng Lyu$^{4}$ and I-Ming Chen$^{3}$\\[4pt]
\small $^{1}$The Hong Kong University of Science and Technology; $^{2}$ZENBOT; $^{3}$Nanyang Technological University\\
\small $^{4}$The Hong Kong Polytechnic University\qquad $^{*}$Correspondence: \href{mailto:eeguan@ust.hk}{eeguan@ust.hk}}
\date{}

\begin{document}
\maketitle

\begin{abstract}
\noindent Connectome models of the fly nerve cord generate walking-like motor rhythms, but oscillation alone does not show that the specific wiring matters. Here we provide, to our knowledge, the first test of which features of motor output depend on the specific wiring. We simulated the leg motor systems of two independent \textit{Drosophila} connectomes, with synapse counts as fixed weights and glutamatergic synapses treated as inhibitory, and compared each with six families of rewired networks that preserve progressively more of its structure, using pre-registered criteria. We find that rhythm is generic but antagonist coordination is not: many rewired networks were more rhythmic than the real ones, yet the real wiring coordinated antagonistic motor pools more strongly than every rewired network, most of all at the thorax--coxa joint. We trace this specificity to how premotor input is allocated between antagonistic pools. Both connectomes carry Sherrington's reciprocal innervation---neurons that excite one pool inhibit its antagonist---and no rewired network does. Reassigning premotor inputs between the pools abolished coordination even when motor neurons' typical input strength changed little (all pre-registered criteria met in one connectome; same direction in the other). Coordination, not rhythm, therefore reveals whether a connectome model's wiring matters.
\end{abstract}

\section*{Introduction}
\begin{figure}[tbp]
\mainfig{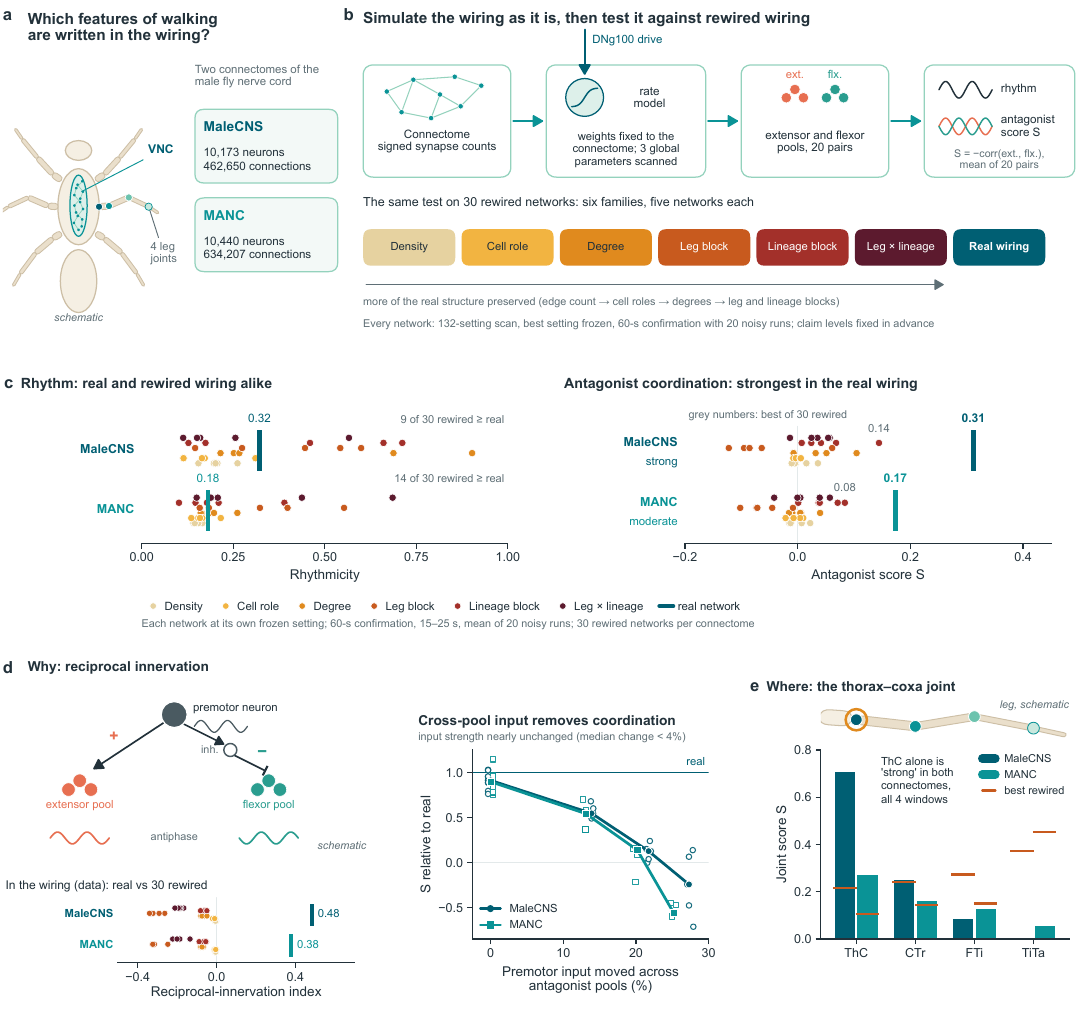}
\caption{\textbf{\capTitleA} \capBodyA}
\label{fig:overview}
\end{figure}
How much of behaviour is written in the wiring? Complete synapse-resolution wiring diagrams of the adult fly nerve cord, from the male adult nerve cord (MANC)\citep{takemura2024manc,marin2024,cheong2026}, the female adult nerve cord (FANC)\citep{azevedo2024fanc} and the complete male central nervous system (MaleCNS)\citep{malecns}, make this question concrete, and connectome-constrained simulations now reproduce sensorimotor transformations\citep{shiu2024}, visual responses\citep{lappalainen2024} and rhythmic leg motor activity\citep{pugliese2025}. Here we answer it for fly walking in connectome models (Fig.~\ref{fig:overview}): the leg motor rhythm does not require the real wiring, whereas the real wiring coordinates antagonistic motor pools more strongly than any rewired network and carries Sherrington's reciprocal innervation.

A model that produces a behaviour does not, however, show that its specific wiring matters. With learned interfaces, even the connectome of a nematode can drive realistic walking of a simulated fly body\citep{sphinx2026}, and a whole-brain fly connectome used as the graph of a controller trained by reinforcement learning can control a simulated fly body\citep{flygm2026}. Recurrent networks oscillate readily, and in connectome-derived networks degree and weight statistics fix much of the gross dynamics: in a frozen rate model of the larval connectome, degree- and weight-matched rewiring reproduced the gross dynamical properties, whereas input routing required the exact wiring\citep{therianos2026}, and apparent advantages of connectome topology in trained networks largely disappeared under degree-preserving controls\citep{dhiman2026}. Testing whether wiring is specific therefore requires anatomical interfaces without learned components, and controls that keep generic structure while scrambling specific connections.

Walking has at least two separable components: rhythm generation, and pattern formation, which includes the alternation of antagonistic muscles\citep{mccrea2008}. More than a century ago, Sherrington attributed antagonist alternation to reciprocal innervation, in which pathways that excite one muscle inhibit its antagonist\citep{sherrington1906}. In the fly, leg premotor networks cluster into modules that link motor neurons of muscles with related functions\citep{lesser2024}; GABAergic premotor neurons of the 13A and 13B hemilineages inhibit specific groups of motor neurons, disinhibit their antagonists and can drive flexion--extension alternation during grooming\citep{syed2026}; and connectome simulations identified a three-neuron circuit, driven by DNg100, that was necessary and sufficient for leg motor rhythms in four connectome datasets, with a consistent phase offset between antagonistic coxa promotor and remotor motor neurons within a leg\citep{pugliese2025}. Whether such antagonist coordination requires the specific wiring, beyond what generic network structure provides, has, to our knowledge, not been tested.

We therefore test, with nested null models and pre-registered criteria, which features of leg motor output are written in the wiring, in fixed-weight models of two independent connectomes. We find that rhythm is everywhere but that the real wiring coordinates antagonists more strongly than any rewired network; that the real wiring carries Sherrington's reciprocal innervation; that reallocating premotor input between antagonistic pools abolishes coordination even when motor neurons' input strength changes little (median change 2--4\%); and that the coordination is concentrated at the thorax--coxa joint and persists for minutes.

\section*{Results}

\subsection*{Fixed-weight models of the leg motor system from two independent connectomes}
We built models of the leg motor system from two independently reconstructed connectomes, MaleCNS and MANC, with connection weights fixed at signed synapse counts (Supplementary Fig.~\ref{edfig:design}a,b). Each subnetwork comprised the leg motor neurons, the proprioceptive sensory neurons of the leg nerves, and the descending and intrinsic neurons with at least half of their nerve-cord synapses in the leg neuropils: 10,173 neurons (8,702 interneurons, 424 descending neurons, 381 motor neurons and 666 sensory neurons) and 462,650 connections of at least five synapses in MaleCNS, and 10,440 neurons (8,951, 498, 396 and 595, respectively) and 634,207 connections in MANC. No connection weight was trained or fitted, and glutamatergic synapses were treated as inhibitory, as in previous whole-brain models\citep{shiu2024}. The interfaces were anatomical: input came only through the two DNg100 neurons, descending neurons whose activation initiates forward walking\citep{sapkal2024,pugliese2025}, and output was read only from motor neurons, grouped by annotated target muscle into extensor and flexor pools at the four joints of each leg: thorax--coxa (ThC), coxa--trochanter (CTr), femur--tibia (FTi) and tibia--tarsus (TiTa). Each connectome yielded 20 antagonistic pool pairs (six each at ThC, CTr and FTi and two at TiTa).

Dynamics followed the rate formulation of Pugliese et al.\citep{pugliese2025}, simplified to identical units (Methods):
\begin{equation}
\tau\,\dot{\mathbf r}=-\mathbf r+\big[\tanh\big(g\,(\mathbf W_E+\beta\,\mathbf W_I)\,\mathbf r+\mathbf I+\boldsymbol\xi\big)\big]_+,
\label{eq:model}
\end{equation}
where $\mathbf r$ holds the rates of all neurons, $\mathbf W_E$ and $\mathbf W_I$ the signed synapse counts from excitatory and inhibitory neurons, $\mathbf I$ a constant drive $d$ to the two DNg100 neurons, $\boldsymbol\xi$ weak input noise, $[x]_+=\max(x,0)$ and $\tau=20$~ms. Only three global parameters were free, the gain $g$, the inhibition scale $\beta$ and the drive $d$, and for every network they were selected by the same scan of the antagonist score over 132 settings.

Our primary readout, the antagonist score, measures how strongly antagonistic pools alternate. For an antagonistic pair $p$, let $e_p(t)$ and $f_p(t)$ be the mean demeaned rates of its extensor and flexor pools, and let the amplitude gate $G_p$ be 1 when both are substantially modulated and 0 otherwise (Methods). Over a set $\mathcal A$ of pairs,
\begin{equation}
S_{\mathcal A}=\frac{1}{|\mathcal A|}\sum_{p\in\mathcal A}G_p\,\big[-\rho(e_p,f_p)\big],
\label{eq:S}
\end{equation}
where $\rho$ is the Pearson correlation over the scoring window; $S$ denotes the score over all 20 pairs. $S$ is thus positive when antagonists alternate and negative when they are co-active. Rhythmicity was measured separately, as the spectral concentration of individual motor-neuron signals (Methods).

We compared each real network with six families of rewired networks, five networks each, that preserve progressively more of its structure: the number of edges (density); the number of edges between cell roles (cell role); in addition, each neuron's in- and out-degree (degree); and edge counts between blocks defined by leg segment and side (leg block), by developmental hemilineage (lineage block) or by both (leg $\times$ lineage) (Supplementary Fig.~\ref{edfig:design}c). Every neuron kept its transmitter sign. Three features of the design are rare in connectome modelling. The claim levels and decision criteria were recorded before the corresponding results were seen (Methods). Every comparison was confirmed with fresh-noise 60-s simulations of each network at its frozen best setting, with 20 noise realizations and three noise-free initial states, scored in four windows (Supplementary Fig.~\ref{edfig:design}d). And the confirmation runs regenerate exactly: an independent second re-simulation reproduced all 4,692 run-window records of each connectome to within rounding error.

\subsection*{Rhythm is everywhere; antagonist coordination is strongest in the real wiring}
\begin{figure}[tbp]
\mainfig{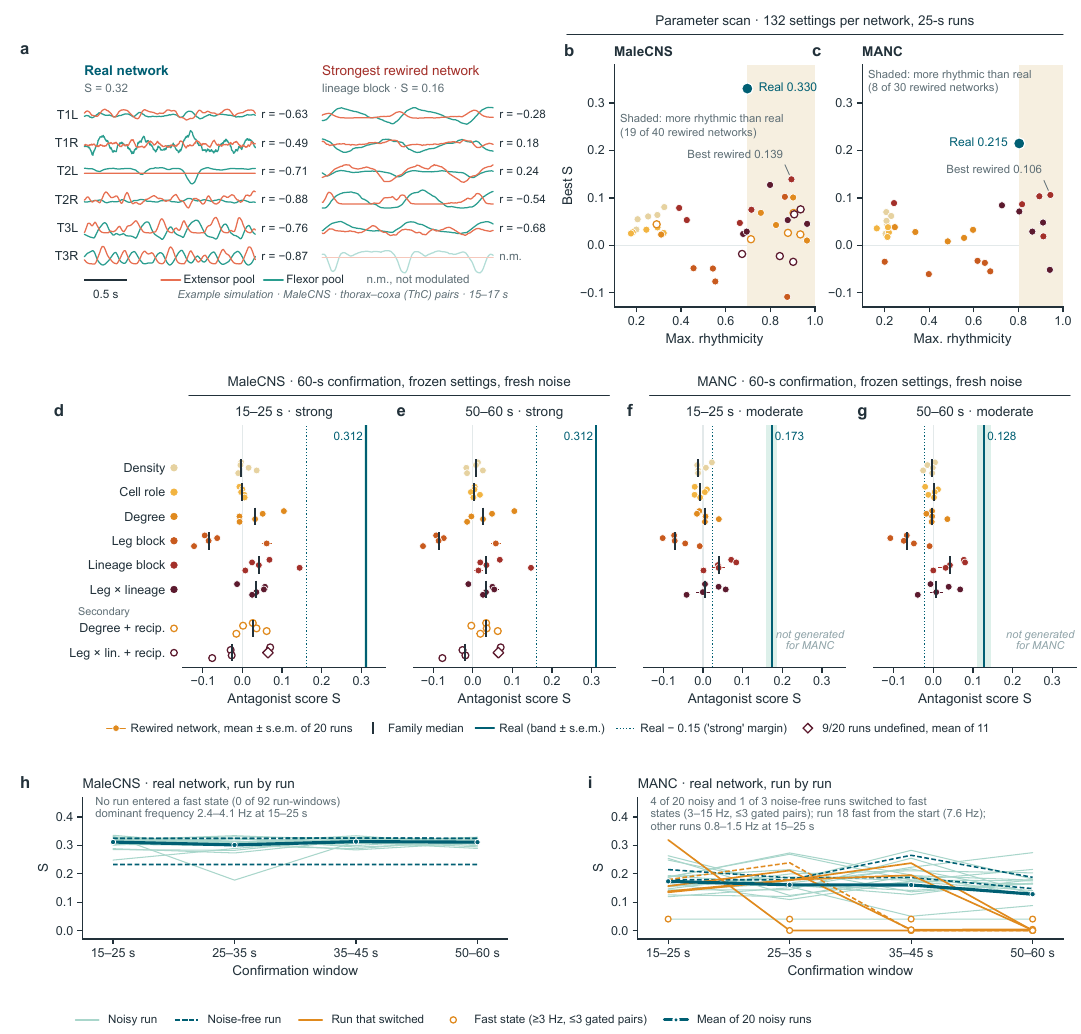}
\caption{\textbf{\capTitleB} \capBodyB}
\label{fig:main}
\end{figure}

Rewired networks oscillated as readily as the real networks, and many were more rhythmic. In the parameter scan, 19 of 40 rewired MaleCNS networks and 8 of 30 rewired MANC networks reached a higher maximum rhythmicity than the corresponding real network (MaleCNS: up to 0.96 versus 0.70; Fig.~\ref{fig:main}b,c). At the settings selected for coordination, rewired MaleCNS networks reached a rhythmicity of up to 0.90 in the confirmation runs, against 0.32 for the real network.

The real wiring drove antagonistic pools into antiphase more strongly than any rewired network (Fig.~\ref{fig:main}a--g). In the confirmation runs, the real MaleCNS network scored 0.312 in both main windows, against at most 0.144 and 0.147 for any of its 40 rewired networks, and the real MANC network scored 0.173 and 0.128, against at most 0.084 and 0.080. The real network exceeded every rewired network in every confirmation window of both connectomes, including two further MaleCNS families that also preserve the number of reciprocally connected neuron pairs (Fig.~\ref{fig:main}d,e, hollow symbols). By the pre-registered criteria, MaleCNS reached the strong level in all four windows, with margins of 0.26--0.40 over every family median, and MANC the moderate level in all four windows, its margin over the lineage-block family median (0.134 and 0.086 in the two main windows) falling short of the 0.15 required for strong (Supplementary Table~\ref{tab:levels}). This meets the pre-registered criterion for replication in a second connectome. The scan had given the same picture (best real scores 0.330 and 0.215, against at most 0.139 and 0.106 for rewired networks; Fig.~\ref{fig:main}b,c). Coordination did not require noise: in MaleCNS, the three noise-free runs settled at 0.325, 0.233 and 0.325 and stayed there for 60~s.

\subsection*{The real wiring carries Sherrington's reciprocal innervation}
\begin{figure}[tbp]
\mainfig{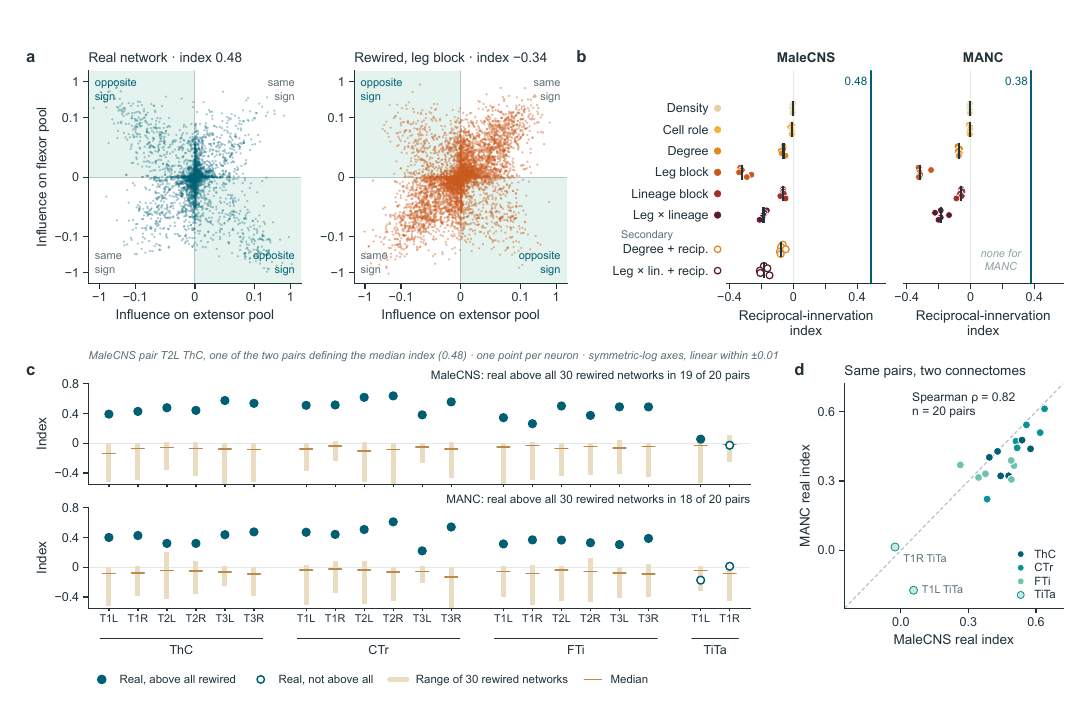}
\caption{\textbf{\capTitleC} \capBodyC}
\label{fig:struct}
\end{figure}

Both real networks, and none of the rewired networks, carry the structural signature of reciprocal innervation. For each antagonistic pair we computed every neuron's signed one- and two-step influence on the extensor pool and on the flexor pool; the reciprocal-innervation index is minus the correlation of these two influences across neurons and is positive when neurons that drive one pool suppress its antagonist (Fig.~\ref{fig:struct}a; Methods). The index was 0.48 in MaleCNS and 0.38 in MANC (median over the 20 pairs) and negative in every rewired network (Fig.~\ref{fig:struct}b). It exceeded all 30 rewired networks at 19 of 20 pairs in MaleCNS and 18 of 20 in MANC, the exceptions being tibia--tarsus pairs (Fig.~\ref{fig:struct}c), and its profile across pairs was shared by the two connectomes (Spearman $\rho=0.82$; Fig.~\ref{fig:struct}d).

The index reflects opposite-signed influence rather than segregated targeting (post hoc; Methods). Of the influence on antagonistic pools, 96\% (MaleCNS) and 97\% (MANC) came from neurons that influence both pools, and 88\% and 83\% of this shared influence was opposite in sign, against 16--49\% in the 30 rewired networks (medians over pairs); the real value exceeded all 30 rewired networks at 18 of 20 pairs in both connectomes (Supplementary Table~\ref{tab:fopp}). Because a neuron's direct outputs share its transmitter sign, opposite-signed influence arises only through an intervening neuron: from excitatory neurons that excite one pool and, through an inhibitory interneuron, inhibit its antagonist (45\% and 42\% of the opposite-signed influence), and from inhibitory neurons that inhibit one pool and disinhibit its antagonist (55\% and 58\%).

Without the specific wiring, antagonistic pools of the same leg receive input of the same sign from the same neurons: rewiring within leg blocks made the index strongly negative (Fig.~\ref{fig:struct}a,b). The index was high also at pairs that were not active in the simulations (compare Fig.~\ref{fig:struct}c with Fig.~\ref{fig:joints}a): reciprocal innervation is a network-wide organization of the wiring, which the dynamics express where the pools are recruited.

\subsection*{Reallocating premotor input between antagonistic pools abolishes coordination, even with input strength nearly unchanged}
\begin{figure}[tbp]
\mainfig{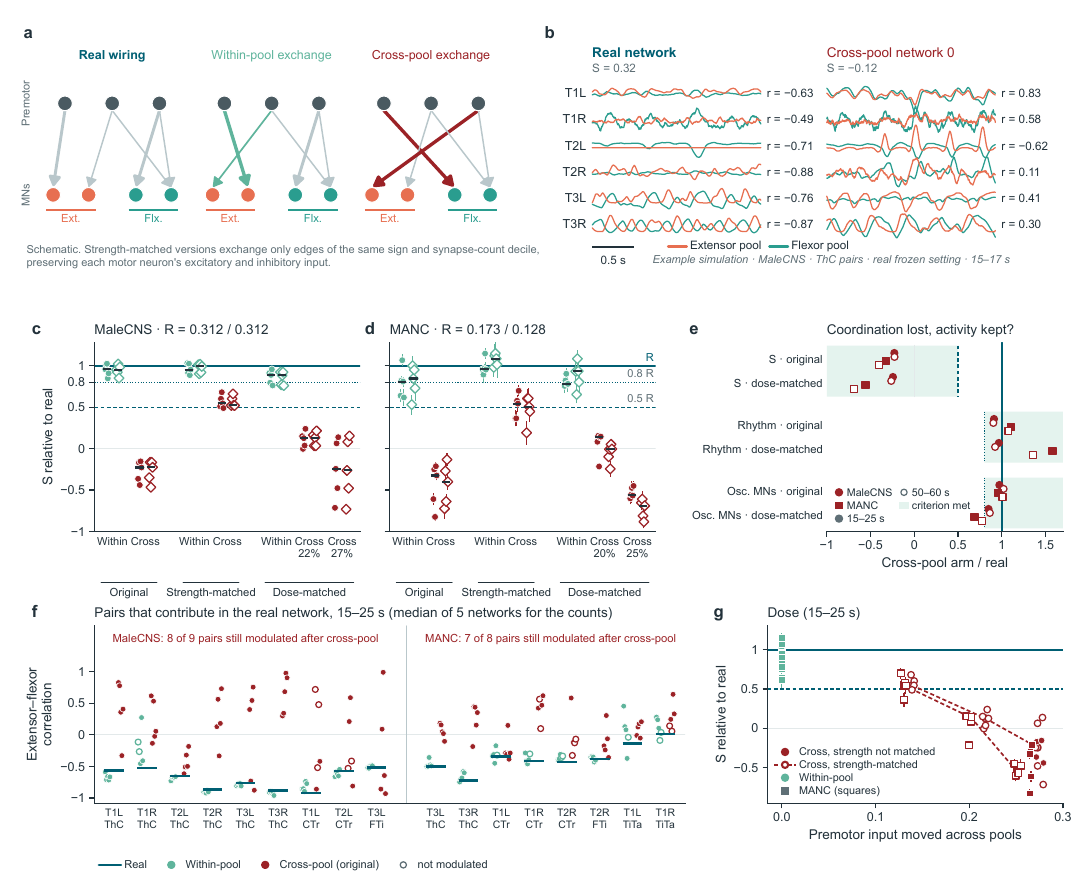}
\caption{\textbf{\capTitleD} \capBodyD}
\label{fig:surgery}
\end{figure}

Reassigning premotor inputs across antagonistic pools abolished coordination while the pools stayed active. We exchanged the targets of edges onto motor neurons, either between the extensor and flexor pools of the same leg and joint (cross-pool) or within a pool (within-pool control), keeping every motor neuron's number of inputs and every presynaptic neuron's outputs (Fig.~\ref{fig:surgery}a). At the real network's frozen setting, cross-pool reassignment drove $S$ below zero in both connectomes (median of five networks: MaleCNS $-0.07$ in both main windows; MANC $-0.06$ and $-0.05$), so that antagonistic pools became co-active, whereas within-pool controls kept 81--96\% of the real score (Fig.~\ref{fig:surgery}b--d). Rhythmicity and the fraction of oscillating motor neurons were preserved, and 8 of the 9 contributing pairs in MaleCNS and 7 of the 8 in MANC stayed modulated on both sides (Fig.~\ref{fig:surgery}e,f). The outcome met all six pre-registered conditions for a selective loss of coordination in MaleCNS in all four windows, and in MANC in the primary window, where 96\% of the oscillating motor neurons kept oscillating; in the other three MANC windows, one to three conditions were not met, including pair retention in the secondary window (4 of 6 contributing pairs, against 5 of 6 for the within-pool controls; Supplementary Tables~\ref{tab:surgery} and~\ref{tab:c6}). At each network's own selected setting, MANC met all six conditions in both main windows (condition 6 assessed from the number of modulated pairs), whereas the MaleCNS cross-pool networks, selected at less rhythmic settings, failed the rhythm condition.

Coordination collapsed even when motor neurons' excitatory and inhibitory input changed little. Exchanging edges of different synapse counts also changes how much excitation and inhibition each motor neuron receives (median relative change 27--34\%). A first strength-matched version, restricted to edges of the same transmitter sign and synapse-count decile (median input change 1.2--1.9\%), reduced coordination only partially (15--25~s: MaleCNS 0.17 versus 0.30 for its within-pool control and 0.31 for the real network; MANC 0.09 versus 0.17 and 0.17). Every strength-matched cross-pool network scored below every control, but the reduction missed the pre-registered magnitude criteria (MaleCNS 0.55 and 0.52 of the real score, where at most half was required), and a structural audit made after these results showed why: the construction had moved only about half as much premotor input across antagonistic pools as the original one (13.9\% versus 27.2\% in MaleCNS; 13.1\% versus 26.5\% in MANC; Fig.~\ref{fig:surgery}g). We therefore designed and pre-registered a dose-matched test before running any of its simulations. Its strength-matched cross-pool arm (\Xhi) moved as much input across pools as the original reassignment (median 27\% in MaleCNS and 25\% in MANC) while changing motor neurons' excitatory and inhibitory input by a median of 2.0--3.9\% (95th percentile 11--30\%; within-pool controls, 1.0--2.3\% and 9--19\%; Supplementary Fig.~\ref{edfig:dose}a). Input changes of this order left coordination largely intact in the within-pool controls, so the loss of coordination described below cannot be attributed to them.

In MaleCNS, the dose-matched reassignment abolished coordination and met every pre-registered condition. $S$ fell to $-0.076$ and $-0.082$ in the two main windows ($-0.24$ and $-0.26$ of the real score), every \Xhi\ network scored below every within-pool control, which kept 0.89 and 0.88 of the real score, and rhythmicity (0.31, against 0.32 for the real network) and activity (8 of 9 contributing pairs; 85\% of oscillating motor neurons) were retained (Fig.~\ref{fig:surgery}c,e; Supplementary Fig.~\ref{edfig:dose}b). All six conditions were also met at 25--35~s; at 35--45~s, cross-pool rhythmicity fell just short of condition 5 (Supplementary Table~\ref{tab:exp2}). By the pre-registered interpretation, coordination in MaleCNS depends on which antagonistic pool premotor input is allocated to, not on changes in the excitation and inhibition that motor neurons receive, which were of similar size in the within-pool controls that kept coordination; this holds at the real network's parameters and does not distinguish allocation by individual neurons from allocation by developmental lineages. The secondary arms agreed in direction. A lower-dose arm (\Xlo) reduced $S$ to 0.13 of the real score but kept 7 of 9 contributing pairs, just short of condition 6. When every network was re-optimised, \Xhi\ networks again lost coordination ($S$ close to 0), but in a different regime, with low rhythmicity and about twice as many oscillating motor neurons as the real network, and with re-optimised controls at 0.73 and 0.77 of the real score, so conditions 1, 5 and 6 were not met there (Supplementary Fig.~\ref{edfig:dose}c,d).

MANC reproduced the collapse of coordination and the pre-registered ordering. After dose-matched reassignment, $S$ fell to $-0.096$ and $-0.088$ ($-0.56$ and $-0.69$ of the real score), all five \Xhi\ networks scored below zero and below every within-pool control, and rhythmicity rose to 1.58 and 1.36 times that of the real network. Motor-neuron activity retention (69--77\% of the real network's oscillating motor neurons, in all four windows) fell short of the pre-set 80\% criterion, and in the primary window the within-pool controls kept 0.78 of the real score, just below the 0.8 required, so the pre-registered test was not met in MANC (Supplementary Fig.~\ref{edfig:dose}b,d). At re-optimised settings, conditions 1--5 were met in both windows and condition 6 again was not. In both connectomes the scores after reassignment were negative: the antagonistic pairs that remained active became co-active, rather than merely falling silent.

Across the strength-matched constructions, coordination fell steadily as more premotor input was moved between antagonistic pools (Fig.~\ref{fig:surgery}g). In MaleCNS, $S$ relative to the real score was 0.55 after moving 13.9\% of the input (original strength-matched arm), 0.13 after 21.7\% (\Xlo), $-0.24$ after 27.3\% (\Xhi) and $-0.23$ after 27.2\% without strength matching; in MANC, 0.54, 0.14, $-0.56$ and $-0.32$. At a similar dose, matching motor neurons' input strength did not preserve coordination. The arms of this series differ in more than dose, because the original strength-matched arm kept a restriction to presynaptic lineage blocks that \Xlo\ and \Xhi\ relax, but the two most direct comparisons point the same way: \Xlo\ and \Xhi\ share one construction, and \Xlo, matched to the within-pool control in the fraction of new edges and similar to it in the change of input strength (median 1.5--2.6\% against 1.0--2.3\%; 95th percentile 9--20\% against 9--19\%), kept only 0.13 and 0.14 of the real score, against 0.89 and 0.78 for the control.

\subsection*{Coordination is concentrated at the thorax--coxa joint and persists for minutes}
\begin{figure}[tbp]
\mainfig{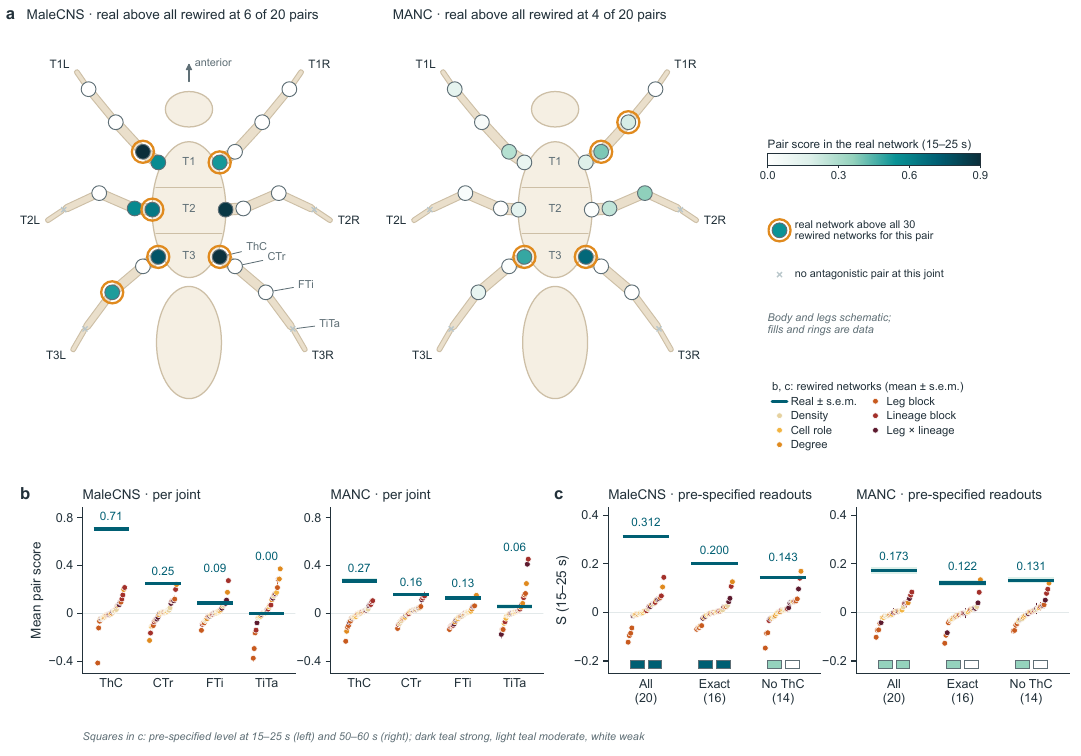}
\caption{\textbf{\capTitleE} \capBodyE}
\label{fig:joints}
\end{figure}

The wiring-specific coordination is concentrated at the thorax--coxa joint. In the real MaleCNS network, 9 of the 20 antagonistic pairs contributed, six of them at the thorax--coxa joint; per joint, the mean pair score was 0.71 at ThC, 0.25 at CTr, 0.09 at FTi and 0 at TiTa, and in MANC 0.27, 0.16, 0.13 and 0.06 (Fig.~\ref{fig:joints}a,b). The thorax--coxa readout was the only one of the seven readouts to reach the strong level in both connectomes and all four windows (primary window 0.71 and 0.27, against at most 0.22 and 0.11 for any rewired network; Supplementary Fig.~\ref{edfig:levels}a). Nearly every antagonistic pair that was active in the real networks alternated: 16 of 17 active pairs were antiphasic, against 48--51\% in the rewired networks (post hoc; Supplementary Note~3). The distal joints scored low because few of their pairs were recruited, not because their pools were in phase.

Two pre-registered control readouts confirmed the advantage in the primary window. Because one pool of each thorax--coxa pair in the middle and hind legs consists of sternal anterior rotator motor neurons whose joint assignment is approximate, we pre-registered $S$ over the 16 pairs formed only from exactly assigned motor neurons and $S$ over the 14 pairs outside the thorax--coxa joint, computed on the confirmation trajectories. In the primary window, both reached at least the moderate level in both connectomes, meeting the pre-registered criterion (Fig.~\ref{fig:joints}c). With exact assignments only, MaleCNS scored 0.200 against at most 0.126 for any rewired network (strong) and MANC 0.122, with one rewired network at 0.134 (moderate). Without the thorax--coxa pairs, MaleCNS scored 0.143, with one rewired network at 0.169, and MANC 0.131 (both moderate). The advantage outside the thorax--coxa joint was weaker later: at 50--60~s, both readouts in MANC and the non-thorax--coxa readout in MaleCNS fell to the weak level (Supplementary Fig.~\ref{edfig:levels}a; Supplementary Table~\ref{tab:readouts}). The thorax--coxa joint therefore carries the robust part of the advantage.

The coordinated state persisted for minutes. In MaleCNS, no run ever left it: none of the 92 run-windows of the 60-s confirmation was in a fast, uncoordinated state, and all 40 noisy and 10 noise-free 300-s trajectories stayed coordinated for the full 5~min (Fig.~\ref{fig:main}h; Supplementary Fig.~\ref{edfig:persist}). The coordinated MANC state was slower (dominant frequency about 1~Hz, against 2.4--4.1~Hz in MaleCNS) and occasionally switched to a fast, uncoordinated rhythm: in the 60-s confirmation, 15 of 20 runs stayed in the slow state throughout, 4 switched abruptly into fast (3.4--14.6~Hz) states with at most three modulated pairs, 1 was in such a state from the first window, and one of the three noise-free runs also switched (Fig.~\ref{fig:main}i). Over 5~min, coordinated episodes of the real MANC network lasted about 4.2~min on average, against about 11--16~s for its two strongest rewired networks (departures from the coordinated state 0.24 per coordinated minute, 95\% confidence interval 0.16--0.35, against 3.7 and 5.6; Supplementary Table~\ref{tab:exp3}). At 290--300~s, 45\% of the real MANC trajectories were coordinated, against 10\% and 0\%, and the ensemble mean of $S$ was 0.088, against 0.076 and 0.072. By the pre-registered rules this is an intermediate outcome: the ordering held, which rules out the pre-registered reading that it holds only on the scale of a minute, but fewer than the 50\% of trajectories required to call the state persistent remained coordinated, and the margin in mean $S$ was small (post hoc bootstrap 95\% interval for the difference from the stronger rewired network, $-0.013$ to 0.036).

\section*{Discussion}
Sherrington's principle of reciprocal innervation, proposed more than a century ago, is written directly into the fly leg connectome, and in models whose connection weights are fixed at synapse counts it emerges in the dynamics as antagonist coordination. Rhythm, by contrast, is generic: rewired networks with the same coarse statistics oscillated as readily as the real ones. In both connectomes the real wiring coordinated antagonists more strongly than every rewired network, and coordination depended on which antagonistic pool each premotor input reaches: reallocating inputs abolished it even when motor neurons' excitatory and inhibitory input changed little, meeting every pre-registered criterion in MaleCNS. These conclusions rest on design features that are rare in connectome modelling: criteria recorded before the results, replication in a second, independently reconstructed connectome, and confirmation runs that regenerate exactly.

For connectome modelling in general, the lesson is that wiring specificity should be tested on coordination, not on rhythm. A connectome model that oscillates passes a test that rewired networks pass too. This complements connectome simulations that identified a three-neuron rhythm-generating circuit downstream of DNg100, together with a within-leg phase offset between coxa promotor and remotor motor neurons\citep{pugliese2025}. Our results do not argue against that circuit, and the real network may well generate its rhythm through it. The two approaches answer different questions: pruning identifies circuits within the real wiring that are sufficient for a behaviour of the model, whereas rewired networks ask which features of the model's output depend on the specific wiring. Antagonist coordination is such a feature, and the thorax--coxa pair at which Pugliese and colleagues observed a phase offset is the joint at which we find the wiring-specific coordination concentrated.

The organization we find agrees with anatomical and functional studies of fly premotor circuits. Premotor modules link motor neurons of muscles with related functions\citep{lesser2024}, and inhibitory 13A and 13B neurons can suppress one group of motor neurons while disinhibiting its antagonists\citep{syed2026}. In the connectome, reciprocal innervation is a property of the network-wide input pattern and takes both forms: excitatory neurons excite one pool and, through inhibitory interneurons, inhibit its antagonist, and inhibitory neurons inhibit one pool and disinhibit its antagonist, the pattern described for 13A and 13B neurons, which carries slightly more of the opposite-signed influence (post hoc). Consistent with a distributed implementation, in earlier simulations under a shorter protocol, cutting the 13A/13B inhibition--disinhibition motif did not reduce coordination more than weight-matched control cuts, and in a knockout screen no single cell type stood out from size-matched random deletions (Supplementary Note~1).

These results concern the model, not the animal, and several limitations apply. They assume that glutamatergic transmission onto these neurons is inhibitory; in earlier short-window simulations, treating glutamate as excitatory removed the real network's advantage over lineage-constrained networks (Supplementary Note~1). They use a single rate-neuron model, synapse counts as weights and predicted transmitter signs, with no body and no sensory feedback; in the animal, coordination within a leg also depends on load and descending inputs\citep{sapkal2026}. About a quarter of the motor-neuron joint assignments are approximate. In MANC, the thorax--coxa advantage comes mainly from hind-leg pairs whose promotor-side pool consists of approximately assigned sternal rotators, whereas in MaleCNS the front-leg pairs formed from exactly assigned motor neurons alone are also specific (Supplementary Table~\ref{tab:thc}). MANC reproduced each main finding in direction, with smaller margins: moderate rather than strong, a coordinated state that switched on a timescale of minutes, and a dose-matched test that fell short of its activity-retention criterion and, in the primary window, of its control criterion. Each rewired family contains five networks (smallest attainable within-family rank $p$, 1/6). Parameters were searched on a finite grid, with the MANC setting at the highest drive, and the pre-registration of the first comparison was only partly blind. The predicted antagonist phase relationships remain to be compared with recordings from walking flies\citep{azevedo2020} and tested in further connectomes\citep{azevedo2024fanc}. Within these bounds, the models indicate that what the fly leg connectome specifies beyond generic network structure is how premotor input is allocated between antagonistic muscles, rather than the capacity to generate rhythm.

\section*{Methods}

\subsection*{Connectome data and subnetwork extraction}
We used the MaleCNS v1.0 connectome\citep{malecns} (synapse confidence $\geq 0.5$) and the MANC v1.0 connectome\citep{takemura2024manc,marin2024,cheong2026}, with the same extraction rules for both. Motor neurons were the leg motor neurons of the three thoracic segments, and sensory neurons were the proprioceptive sensory neurons entering through the leg nerves. Descending neurons and intrinsic (including ascending) neurons were included when at least half of their ventral nerve cord synapses (output synapses for descending neurons) lay in the leg neuropils and they had at least 20 leg-neuropil synapses. Connections were synapse counts within the ventral nerve cord, retained when they had at least five synapses, with self-connections removed. Signs followed the predicted transmitter of the presynaptic neuron: acetylcholine $+1$; GABA and glutamate $-1$ (and, in MaleCNS, histamine $-1$); other transmitters 0. Glutamatergic transmission was therefore treated as inhibitory throughout. In MANC, DNg100 corresponds to the type DNxl058 (two neurons).

\subsection*{Motor-neuron pools}
Motor neurons were mapped from their annotated type to a joint (ThC, CTr, FTi or TiTa) and a direction (extensor-like or flexor-like). Of 381 MaleCNS motor neurons, 302 were mapped, 76 of them approximately; of 396 MANC motor neurons, 302 were mapped, 83 approximately. Approximate assignments were sternal anterior and posterior rotator motor neurons assigned to the promotion--remotion axis of the thorax--coxa joint, and long-tendon-muscle motor neurons assigned to tarsal depression. A pool is the set of motor neurons of one leg, joint and direction, and an antagonistic pair exists when both directions are represented, giving 20 pairs in each connectome. Restricting pools to exactly assigned motor neurons leaves 16 pairs (CTr 6, FTi 6, ThC 2, TiTa 2).

\subsection*{Rate model}
The model follows the rate formulation of Pugliese et al.\citep{pugliese2025} (rectified-tanh units, a mean time constant of 20~ms, weights equal to synapse counts with a five-synapse floor, GABAergic and glutamatergic synapses inhibitory, tonic DNg100 drive), simplified to identical units with three global parameters, so that the real network and every rewired network receive the same parameter search. The two models are not identical: in that model the parameters of every unit, including a threshold and a time constant (mean 20~ms, standard deviation 2~ms), are drawn at random and the equations are solved with an ODE solver, whereas here all units share one time constant, have no threshold and receive a small input noise. Rates follow Eq.~(\ref{eq:model}) with weights
\begin{equation}
w_{ij}=\sigma_j\,n_{ij}\,\mathds{1}[n_{ij}\geq 5],\qquad (\mathbf W_E)_{ij}=w_{ij}\,\mathds{1}[\sigma_j=+1],\qquad (\mathbf W_I)_{ij}=w_{ij}\,\mathds{1}[\sigma_j=-1],
\label{eq:weights}
\end{equation}
where $n_{ij}$ is the number of synapses from neuron $j$ to neuron $i\neq j$ and $\sigma_j$ is the sign of neuron $j$ (so $\mathbf W_I$ has non-positive entries), and with drive
\begin{equation}
I_i=d\;\mathds{1}[i\in\mathrm{DNg100}].
\label{eq:drive}
\end{equation}
The weights were fixed for every network. Equation~(\ref{eq:model}) was integrated with the Euler method, $\mathbf r(t+\Delta t)=\mathbf r(t)+\frac{\Delta t}{\tau}\big(-\mathbf r(t)+[\tanh(\cdot)]_+\big)$, with $\Delta t=1$~ms, $\tau=20$~ms and independent Gaussian input noise $\xi_i(t)$ of standard deviation 0.01 drawn at every step (zero in noise-free runs). Initial rates were drawn uniformly from $[0,0.01)$, and motor-neuron rates were recorded at 500~Hz. The parameter grid comprised 11 logarithmically spaced gains $g$ from $10^{-4}$ to $10^{-1.5}$, inhibition scales $\beta\in\{0.5,1,2\}$ and drives $d\in\{0.5,1,2,4\}$ (132 settings).
\subsection*{Readouts}
Within a scoring window, each motor-neuron trace was demeaned, and a neuron counted as oscillating when its standard deviation exceeded 0.01. For antagonistic pair $p$, with extensor pool $E_p$ and flexor pool $F_p$,
\begin{equation}
e_p(t)=\frac{1}{|E_p|}\sum_{i\in E_p}\big(r_i(t)-\langle r_i\rangle\big),\qquad f_p(t)=\frac{1}{|F_p|}\sum_{i\in F_p}\big(r_i(t)-\langle r_i\rangle\big),
\label{eq:pools}
\end{equation}
where $\langle\cdot\rangle$ is the mean over the window. With $s^e_p$ and $s^f_p$ the standard deviations of $e_p$ and $f_p$ over the window, the amplitude gate is
\begin{equation}
G_p=\mathds{1}\!\left[\min(s^e_p,s^f_p)\geq 0.001\ \ \text{and}\ \ \frac{\min(s^e_p,s^f_p)}{\max(s^e_p,s^f_p)}\geq 0.2\right],
\label{eq:gate}
\end{equation}
and the antagonist score is Eq.~(\ref{eq:S}). $S$ is taken over all 20 pairs; the two second-round decision readouts use the same equation over the 16 pairs formed only from exactly assigned motor neurons and over the 14 pairs outside the thorax--coxa joint, and the per-joint readouts over the pairs of one joint. $S$ was undefined in the rare runs in which no motor neuron oscillated. For each oscillating motor neuron $i$, let $P_i(\nu)$ be the power spectrum of its demeaned, Hann-windowed trace and $\nu_i^*$ its peak frequency in the 0.5--50-Hz band. Rhythmicity is
\begin{equation}
Q=\operatorname*{median}_i\ \frac{\sum_{|\nu-\nu_i^*|\leq\delta\nu}P_i(\nu)}{\sum_{0.5\leq\nu\leq 50}P_i(\nu)},
\label{eq:rhythm}
\end{equation}
with sums over frequency bins, $\delta\nu$ one bin and the median over oscillating neurons; the dominant frequency is the median of $\nu_i^*$. A setting was eligible when more than 20\% of motor neurons oscillated and their dominant frequency lay between 1 and 50~Hz.
\subsection*{Rewired networks}
Rewired networks preserved: the number of edges (density); the number of edges between each pair of the four cell roles, sensory, descending, interneuron and motor (cell role); in addition every neuron's in- and out-degree within role blocks (degree); edge counts between blocks defined by role, leg segment and side (leg block); by role and, for interneurons, hemilineage (lineage block); or by both (leg $\times$ lineage). Density and cell-role networks placed edges at random, within role blocks for the latter, and their synapse counts were randomly reassigned to the new edges (within role blocks for the cell-role family). The other four families were generated by degree-preserving exchanges of edge targets within blocks\citep{maslov2002}, with ten attempted exchanges per edge; synapse counts travelled with their edges, so every neuron also kept its output strength. In every family, signs were properties of neurons, so each neuron kept its transmitter sign. Five networks were generated per family. For MaleCNS we also generated two secondary families (degree and leg $\times$ lineage) that additionally preserve the number of reciprocally connected neuron pairs, by exchanging unidirectional edges and reciprocal pairs separately.

\subsection*{Selection and confirmation}
Each network was simulated for 25~s at all 132 settings and scored on 15--25~s, and its eligible setting with the highest $S$ was frozen (MaleCNS real network: $g=10^{-2.5}$, $\beta=0.5$, $d=0.5$; MANC real network: $g=10^{-1.75}$, $\beta=1$, $d=4$). The confirmation simulated each frozen setting for 60~s with 20 noise realizations independent of the scan and three noise-free initial states, and scored windows of 15--25, 25--35, 35--45 and 50--60~s; 15--25~s was the primary and 50--60~s the secondary window. A network's value is the mean over its 20 stochastic runs, excluding runs in which $S$ was undefined, and a family's value is the median over its five networks; runs were never treated as independent networks. Undefined runs occurred only in one network of a secondary MaleCNS family (9 of 20 runs in every window; mean 0.064 over defined runs, 0.035 with undefined runs counted as zero), which affects no pre-specified comparison. We classified the fast states of Fig.~\ref{fig:main}h,i descriptively as windows with a dominant frequency of at least 3~Hz and at most three modulated pairs.

\subsection*{Pre-specified criteria}
For a readout, let $S_{\mathrm{real}}$ be the value of the real network, $S_{k,n}$ that of network $n=1,\dots,5$ of family $k=1,\dots,6$, and $M=\max_k\operatorname{median}_nS_{k,n}$. The claim levels were
\begin{equation}
\begin{aligned}
\textit{strong}:&\quad S_{\mathrm{real}}>\max_{k,n}S_{k,n}\ \ \text{and}\ \ S_{\mathrm{real}}-M\geq 0.15,\\
\textit{moderate}:&\quad \#\bigl\{k:S_{\mathrm{real}}>\max_nS_{k,n}\bigr\}\geq 5\ \ \text{and}\ \ S_{\mathrm{real}}>M,
\end{aligned}
\label{eq:levels}
\end{equation}
and \emph{weak} otherwise; that is, strong requires the real network to exceed every network of all six families and every family median by at least 0.15, and moderate to exceed every network in at least five families and every family median. Replication required the second connectome to reach at least the moderate level. These rules were recorded before the scan results for the degree, leg-block, lineage-block and leg $\times$ lineage families were seen; the real MaleCNS network, the density family and three networks of the cell-role family had already been scanned, so the pre-specification was only partly blind. Analysis windows, gates and eligibility were not changed after results were seen. The criteria for the reassignment experiments and for all second-round analyses were recorded before any of the corresponding simulations were run, and no threshold was changed after the results (Supplementary Note~2).

\subsection*{Motor-input reassignment}
Only edges onto motor neurons were modified, by degree-preserving target exchanges (20 attempts per edge; five networks per type), so every motor neuron kept its number of inputs and every presynaptic neuron its outputs. Exchanges were allowed within blocks defined by the presynaptic neuron's role, leg segment, side and hemilineage and by the motor neuron's leg and joint (cross-pool) or leg, joint and direction (within-pool). The strength-matched versions further restricted exchanges to edges of the same presynaptic sign and synapse-count decile. The reassigned networks were evaluated with the same confirmation protocol, both at the real network's frozen setting and at their own selected settings. The net cross-pool redistribution is
\begin{equation}
D=\frac{\tfrac12\sum_j\sum_q\bigl|a'_{jq}-a_{jq}\bigr|}{\sum_j\sum_q a_{jq}},
\label{eq:D}
\end{equation}
where $a_{jq}$ and $a'_{jq}$ are the numbers of synapses that presynaptic neuron $j$ makes onto the motor neurons of pool $q$ (one leg, joint and direction) before and after reassignment, over presynaptic neurons of non-zero sign and assigned motor neurons: half the summed absolute change of each presynaptic neuron's allocation to leg--joint--direction pools, normalized by the total weight onto these pools. Before any of these simulations, we fixed six conditions for a selective loss. With $R$ the score of the real network, $W$ and $X$ the medians over the five within-pool and the five cross-pool networks, $W_n$ and $X_n$ the individual networks and $Q_R$, $Q_W$ and $Q_X$ the corresponding rhythmicities, the conditions were
\begin{equation}
\begin{gathered}
\begin{aligned}
&(1)\quad W\geq 0.8R; && (2)\quad X\leq 0.5R\ \ \text{and}\ \ R-X\geq 0.10;\\
&(3)\quad X\leq 0.5W\ \ \text{and}\ \ W-X\geq 0.10; && (4)\quad \max_nX_n<R\ \ \text{and}\ \ \max_nX_n<\min_nW_n;\\
&(5)\quad Q_X\geq 0.8\,Q_R\ \ \text{and}\ \ Q_X\geq 0.8\,Q_W; &&
\end{aligned}\\
(6)\quad \text{pair retention}\geq 0.8\ \ \text{and}\ \ \text{oscillation retention}\geq 0.8,
\end{gathered}
\label{eq:conditions}
\end{equation}
that is: within-pool controls keep at least 80\% of the real score; the cross-pool median falls to at most half of the real and of the control score, by at least 0.10; every cross-pool network lies below the real network and below every within-pool network; cross-pool rhythmicity is at least 80\% of that of both references; and at least 80\% of the antagonistic pairs that contribute in the real network, and of its oscillating motor neurons, are retained. Because the first confirmation stored only the number of modulated pairs, condition 6 was initially assessed from that number and then, in the pair-level re-analysis, by pair identity: a pair contributes when it passes the amplitude gate in at least half of the real network's runs, and is retained when it does so in at least half of a reassigned network's runs (median over five networks).

The dose-matched test used three arms per connectome with five networks each: a dose-matched cross-pool arm (\Xhi), in which exchanges were restricted to blocks of presynaptic sign, synapse-count decile, leg and joint, without the presynaptic-group restriction, with 2.5 (MaleCNS) or 3.0 (MANC) attempted exchanges per edge (designed median redistribution 0.273 and 0.252); a lower-dose cross-pool arm (\Xlo) with one attempt per edge (0.217 and 0.202); and a within-pool control (W) with the same blocks plus direction. The arms were chosen by CPU-only construction before any simulation, under the pre-specified requirement that the median relative change of motor neurons' excitatory and of their inhibitory input be at most 3.5\% (median over the five networks of an arm). The limit constrains the median, not every motor neuron: the constructed networks changed motor neurons' excitatory and inhibitory input by a median of 2.0--3.5\% (MaleCNS) and 2.2--3.9\% (MANC) in \Xhi\ and 1.0--2.3\% in W (per network), with 95th percentiles of 11--20\%, 11--30\% and 9--19\%, and a minority of motor neurons changed more. \Xhi\ and W networks were scanned over the 132 settings and confirmed both at the real network's frozen setting and at their own selected settings; \Xlo\ networks were confirmed at the real setting only. The primary test compared \Xhi\ with W, and the secondary test \Xlo\ with W, by the six conditions, with condition 6 assessed by pair identity. The pre-registered interpretation was that meeting all six conditions shows that coordination depends on the allocation of premotor input across antagonistic pools rather than on input strength, and that a cross-pool median above half of the real score would show that allocation contributes only partly. Because the presynaptic-group restriction was removed, these networks also changed which pools each hemilineage targets, so the test cannot separate neuron-level from lineage-level allocation.

\subsection*{Pair-level re-analysis}
In a second pre-registered round, we re-simulated the confirmation runs of every real, main rewired and reassigned network (the latter at the real frozen setting) with the same frozen settings and noise streams, storing pair-level outputs. In each connectome, the re-run reproduced all 4,692 archived run-window records to within rounding error (largest difference in $S$, $1.1\times10^{-16}$). Readouts computed on these trajectories were $S$ over all 20 pairs (for comparison with the archive), $S$ over the 16 exactly assigned pairs and $S$ over the 14 pairs outside the thorax--coxa joint (decision readouts), and per-joint scores (descriptive); pairs that did not pass the amplitude gate counted as zero. The claim-level rule above was applied to each decision readout, with the level set by the primary window. If both readouts reached at least the moderate level in both connectomes, we would conclude that the advantage does not depend on the thorax--coxa joint or on approximate assignments; a weak level for the non-thorax--coxa readout would restrict the claim to the thorax--coxa joint, and a weak level for the exact-assignment readout would make the assignments a stated premise.

\subsection*{Long simulations}
We simulated the real MANC network, its two strongest rewired networks in the confirmation (both lineage-block networks) and the real MaleCNS network for 300~s at their frozen settings, in 30-s segments that carried the full state and random-number key, with 20 noise realizations continuing the confirmation noise, 20 new ones and ten noise-free initial states. The continued realizations reproduced the first 60~s of the confirmation runs to within rounding error. A 10-s window was classed as coordinated when
\begin{equation}
S\geq 0.10\quad\text{and}\quad\sum_pG_p\geq 6,
\label{eq:coord}
\end{equation}
that is, with at least six modulated pairs (undefined $S$ counted as zero), and a switch was a departure from the coordinated state followed by at least three non-coordinated windows. The pre-specified readouts, on the 40 noisy trajectories, were the ensemble mean of $S$ at 290--300~s compared with the stronger rewired network, the fraction of trajectories still coordinated at 290--300~s and the switch rate per minute spent coordinated, with exact Poisson 95\% confidence intervals; the mean duration of a coordinated episode is the inverse of this rate. The pre-specified interpretation was: if the real network's mean did not exceed that of the stronger rewired network, the MANC ordering holds only on the scale of a minute and the coordinated state is metastable; if at least half of the trajectories remained coordinated and the mean exceeded the rewired network's, the state persists with occasional switches; other outcomes are intermediate.

\subsection*{Reciprocal-innervation index}
For each antagonistic pair $p$ and each neuron $k$, the influence on a pool $P\in\{E_p,F_p\}$ combines one- and two-step paths through the signed synapse-count matrix $\mathbf W=(w_{ij})$ of Eq.~(\ref{eq:weights}):
\begin{equation}
u^{(P)}_k=\frac{\bar w^{(1)}_{Pk}}{\max_{k'}\bigl|\bar w^{(1)}_{Pk'}\bigr|}+\frac{\bar w^{(2)}_{Pk}}{\max_{k'}\bigl|\bar w^{(2)}_{Pk'}\bigr|},
\label{eq:influence}
\end{equation}
where $\bar w^{(1)}_{Pk}=|P|^{-1}\sum_{i\in P}w_{ik}$ and $\bar w^{(2)}_{Pk}=|P|^{-1}\sum_{i\in P}(\mathbf W^2)_{ik}$ are the mean signed one-step and two-step weights from $k$ onto the pool's motor neurons, and the maxima run over all neurons. The index of the pair is
\begin{equation}
\mathrm{RI}_p=-\operatorname{corr}_k\bigl(u^{(E_p)}_k,\,u^{(F_p)}_k\bigr),
\label{eq:ri}
\end{equation}
the negative Pearson correlation over all neurons with non-zero influence on either pool; the network-level index is the median of $\mathrm{RI}_p$ over pairs.

A negative correlation across neurons could also arise if the two pools were driven by different neurons, without any neuron pushing them in opposite directions. We therefore decomposed the index (post hoc). With $B_p$ the set of neurons $k$ that influence both pools of pair $p$ ($u^{(E_p)}_k\neq0$ and $u^{(F_p)}_k\neq0$), the share of influence carried by these neurons and the opposite-signed share of their influence are
\begin{equation}
f^{\mathrm{both}}_p=\frac{\sum_{k\in B_p}\bigl(|u^{(E_p)}_k|+|u^{(F_p)}_k|\bigr)}{\sum_{k}\bigl(|u^{(E_p)}_k|+|u^{(F_p)}_k|\bigr)},
\label{eq:fboth}
\end{equation}
\begin{equation}
f^{\mathrm{opp}}_p=\frac{\sum_{k\in B_p}\bigl|u^{(E_p)}_k\,u^{(F_p)}_k\bigr|\;\mathds{1}\bigl[u^{(E_p)}_k\,u^{(F_p)}_k<0\bigr]}{\sum_{k\in B_p}\bigl|u^{(E_p)}_k\,u^{(F_p)}_k\bigr|},
\label{eq:fopp}
\end{equation}
with network-level values the medians over pairs. $f^{\mathrm{opp}}$ is high when the neurons that drive both pools push them in opposite directions. With one-step influence alone, $f^{\mathrm{opp}}=0$ in every network, because a neuron's outputs share its sign. For the real networks we also split the opposite-signed influence (the numerator of Eq.~(\ref{eq:fopp})) by the transmitter sign of neuron $k$. The recomputed index reproduced the archived values to within rounding error in all 31 networks of each connectome.
\subsection*{Post hoc analyses}
The following analyses were made after the corresponding results and are labelled post hoc wherever they appear: the structural audit of net cross-pool redistribution; a bootstrap over trajectories (20,000 resamples) of the difference in mean $S$ at 290--300~s; the decomposition of the reciprocal-innervation index (Eqs.~(\ref{eq:fboth}) and~(\ref{eq:fopp}); Supplementary Table~\ref{tab:fopp}); and two analyses of individual antagonistic pairs on the pair-level confirmation trajectories. For the first of these, a pair counted as active when both of its pools passed the amplitude gate in at least half of a network's 20 runs, and as antiphasic when its mean extensor--flexor correlation over those runs was negative; rewired networks were pooled over the 30 main networks (Supplementary Note~3). For the second, we compared the real network's score at each thorax--coxa pair with those of the 30 main rewired networks, using either all assigned motor neurons or, where possible (front legs only), exactly assigned motor neurons alone (Supplementary Table~\ref{tab:thc}).

\subsection*{Example simulations}
The example traces in Fig.~\ref{fig:main}a and Fig.~\ref{fig:surgery}b come from a CPU re-implementation of the same model with independent (NumPy) noise, not from the confirmation runs. The six thorax--coxa pairs were chosen for display before the traces were viewed. The $S$ of each example run lies within, or next to, the range of the 20 confirmation runs of the same network.

\subsection*{Software}
Simulations used JAX on a single GPU (NVIDIA GeForce RTX 5090); analyses used Python with NumPy, SciPy, pandas and Polars, and figures were drawn with Matplotlib.

\subsection*{Use of large language models}
Large language model assistants (Anthropic Claude and OpenAI ChatGPT and Codex) were used to write and review the analysis, simulation and plotting code, to run simulations under the direction of I.G. and to draft and edit the text. All figures were drawn by plotting code from the simulation outputs; no generative image model was used. All reported numbers were recomputed from the raw simulation outputs by scripts intended for release with the data. The authors reviewed all content and take full responsibility for it.

\section*{Data availability}
The MaleCNS v1.0 and MANC v1.0 connectomes are publicly available from their providers\citep{malecns,takemura2024manc}. The extracted subnetworks, rewired and reassigned networks, simulation outputs, analysis tables, figure source data, dated pre-registration documents and a table tracing every number in this manuscript to its source file will be deposited in a public repository upon publication.

\section*{Code availability}
The custom code for subnetwork extraction, rewiring, simulation, analysis and figures will be deposited in a public repository upon publication.

\clearpage
\setcounter{page}{1}
\renewcommand{\thepage}{S\arabic{page}}
\begin{center}
{\Large\textbf{Supplementary Information}}\\[6pt]
How much of fly walking is written in the wiring?\\[3pt]
Isabel Guan, Yuntian Zhao, Dingyuan Zhang, Shipeng Lyu and I-Ming Chen\\[6pt]
\small Supplementary Notes 1--3 $\cdot$ Supplementary Figures 1--4 $\cdot$ Supplementary Tables 1--9
\end{center}

\subsection*{Supplementary Note 1: Earlier analyses under a superseded short-window protocol}
The analyses in this note used an earlier protocol in which every network was scored on 5-s simulations (1--5~s after onset), with each network's best of 132 settings. They are reported for completeness and as context only: they were not repeated under the protocol of the main text, and none of them is used as evidence for the main conclusions. In this protocol, alternation was also summarized by a correlation-based antiphase index (minus the median extensor--flexor correlation over pairs with both pools active), alongside the amplitude-gated score $S$ of the main text.

\textit{Why the protocol was changed.} In held-out 25-s runs at the setting with the real MaleCNS network's highest 5-s score, the score fell from 0.263 in the early window (1--13~s) to 0.051 in the late window (13--25~s) in essentially every noise realization, whereas rewired and reassigned networks were stationary. The highest 5-s score therefore reflected a transient of about 12~s at high gain. We then adopted the stationary protocol of the main text (25-s runs scored on 15--25~s, followed by 60-s confirmation).

\textit{Glutamate sign.} With glutamatergic synapses treated as excitatory, the real MaleCNS network's antiphase index fell from 0.65 to 0.14 and no longer exceeded the lineage-block and leg $\times$ lineage networks (maxima 0.18 and 0.40). In single deterministic runs, making only the glutamatergic edges onto motor neurons excitatory (5,547 edges) reduced the antiphase index to 0.30 and the gated score to 0.11 (from 0.65 and 0.34) while rhythmicity was kept (0.81), whereas flipping the same number of glutamatergic edges onto other targets, matched in synapse count, did not (antiphase 0.55--0.74 over five controls). Making GABAergic edges onto motor neurons excitatory also reduced them (antiphase 0.14, gated score 0.07, rhythmicity 0.84). In this protocol, inhibitory transmission onto motor neurons thus appeared necessary for the coordination, as reciprocal innervation would predict; all conclusions of the main text are conditional on it.

\textit{Hemilineage activation benchmark (pre-registered, failed).} Before any simulation, we recorded the published extension or flexion phenotype of 12 hemilineages after thermogenetic or optogenetic activation\citep{soffers2025,harris2015} and a model readout (sign of the change in extensor-minus-flexor pool rate at the FTi and CTr joints when the hemilineage was activated). The real MaleCNS network predicted 7 of 12 phenotypes (one-sided binomial $p=0.39$); rewired networks predicted 4--10. A post hoc diagnostic showed that a rule based only on the activated hemilineage's transmitter (cholinergic, extension; inhibitory, flexion) predicted 9 of 12. The labels are thus largely predicted by transmitter identity, which every sign-preserving rewired network inherits, so this benchmark cannot test wiring specificity.

\textit{Knockout screens (negative).} Deleting each of 2,138 units (interneuron types, descending neuron types and hemilineages) at the real network's three best settings identified nine interneuron types whose deletion abolished the antiphase index while oscillation persisted. However, 1--3\% of matched random deletions (one interneuron per leg hemineuromere) did so as well, and only 10 of 2,102 cell types fell below the first percentile of the matched random deletions, against about 21 expected by chance: no single cell type stood out from matched random deletions. Among 35 hemilineages, deleting hemilineage 01A lowered the antiphase index more than each of 1,000 size- and composition-matched random deletions ($p=0.001$), but its effect on the amplitude-gated score was not exceptional ($p=0.17$); we regard it as an exploratory candidate only.

\textit{Targeted cuts of the 13A/13B motif.} Removing inhibition--disinhibition connections among 13A and 13B neurons, or from 13A neurons onto motor neurons\citep{syed2026}, left the antiphase index and the gated score within the range of five control cuts that removed the same numbers of inhibitory edges and synapses from other hemilineages.

\textit{More rewired networks.} With 100 networks per family, the real MaleCNS network's gated score exceeded every network of the degree, leg-block, lineage-block, leg $\times$ lineage and degree + reciprocity families (5-s protocol, whose real-network scores can include the transient described above; this comparison has not been repeated under the stationary protocol, in which each family has five networks).

\subsection*{Supplementary Note 2: Timeline of pre-specification and deviations}
\begin{itemize}\setlength{\itemsep}{1pt}
\item 26 September 2026, about 19:00 (Hong Kong time): claim levels and the primary endpoint for the stationary comparison recorded. The real MaleCNS network, the density family and three cell-role networks had already been scanned under the stationary protocol; no other family had.
\item About 22:10 the same day: the six conditions for motor-input reassignment and the 60-s confirmation design recorded, before any stationary reassignment result.
\item 27 September: confirmation runs executed according to the recorded design, with three implementation changes that do not affect the results (deterministic tie-breaking between settings, which was never needed; resumption after an interruption; atomic file writes). The original analysis script counted undefined runs as zero, whereas the reported means exclude them; this affects one secondary network only (Methods).
\item About 20:50 the same day: second-round pre-registration (pair-level re-analysis, dose-matched reassignment, 300-s simulations) recorded before any second-round simulation. About 21:00: a typographical error in the number of exactly assigned pairs (14 instead of 16; the per-joint counts were correct) was corrected before any result. About 21:15: amendment specifying the dose-matched construction, after CPU-only construction tests and before any simulation of it. In this amendment, the limit on the median relative change of motor neurons' input was relaxed from 3\% to 3.5\%, and matching of the fraction of new edges between arms was delegated to a separate lower-dose arm, because the original requirements could not be met simultaneously.
\item 28 September: all second-round simulations completed and analysed with the pre-registered analysis scripts; no threshold was changed.
\item Deviations from the original plan: condition 6 of the reassignment test was first assessed from counts because pair identities had not been stored, and was re-assessed by identity in the second round; a planned main figure on inhibition at the motor-neuron layer is not included because that analysis has not been repeated under the stationary protocol; the planned expansion to 25 networks per family has not been run.
\item Analyses labelled post hoc in the text: the structural audit of cross-pool redistribution, the bootstrap interval of the difference in mean $S$ at 290--300~s, the decomposition of the reciprocal-innervation index (Supplementary Table~\ref{tab:fopp}), and the two analyses of individual antagonistic pairs (Supplementary Note~3 and Supplementary Table~\ref{tab:thc}). The activity of the dose-matched within-pool controls (Supplementary Fig.~\ref{edfig:dose}d) is descriptive.
\end{itemize}

\subsection*{Supplementary Note 3: Active antagonistic pairs alternate in the real networks (post hoc)}
In the real networks, an antagonistic pair that is recruited almost always alternates. In the primary window, all 9 active pairs of MaleCNS and 7 of the 8 active pairs of MANC were antiphasic (16 of 17), whereas in the 30 main rewired networks only 48\% (MaleCNS) and 51\% (MANC) of the active pairs were antiphasic, as expected if their phase were unrelated to the antagonist relationship (Supplementary Table~\ref{tab:r1}). The pattern held in the secondary window (all 9 active pairs of MaleCNS and all 6 of MANC). The single exception, MANC T1R tibia--tarsus, was near zero correlation. The low scores of the distal joints in the real networks (Fig.~\ref{fig:joints}b) therefore reflect how few of their pairs were recruited, not antagonistic pools that were active in phase. This analysis was made after the pre-registered results and does not enter any pre-registered decision.

\clearpage
\setcounter{figure}{0}
\captionsetup[figure]{name=Supplementary Fig.,labelsep=bar}
\renewcommand{\theHfigure}{S\arabic{figure}}

\begin{figure}[p]
\figpage{183mm}{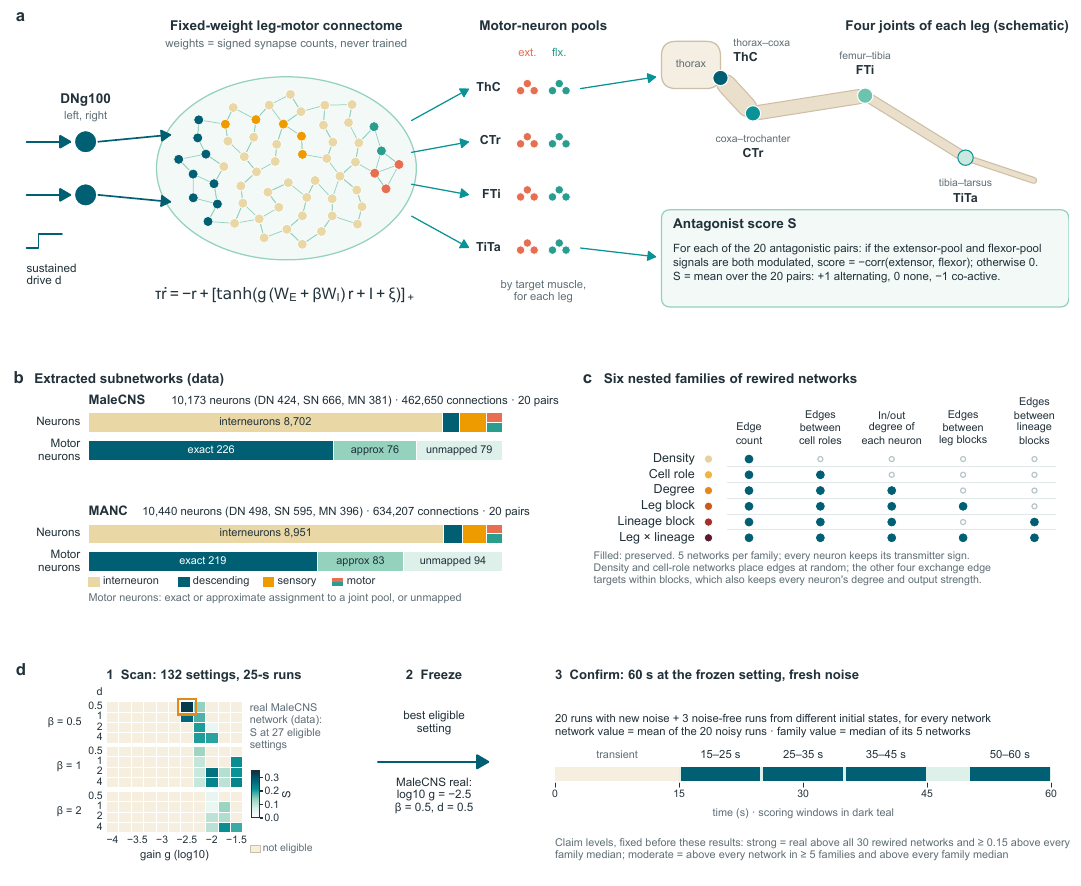}
\caption{\textbf{\capTitleEDA} \capBodyEDA}
\label{edfig:design}
\end{figure}

\begin{figure}[p]
\figpage{183mm}{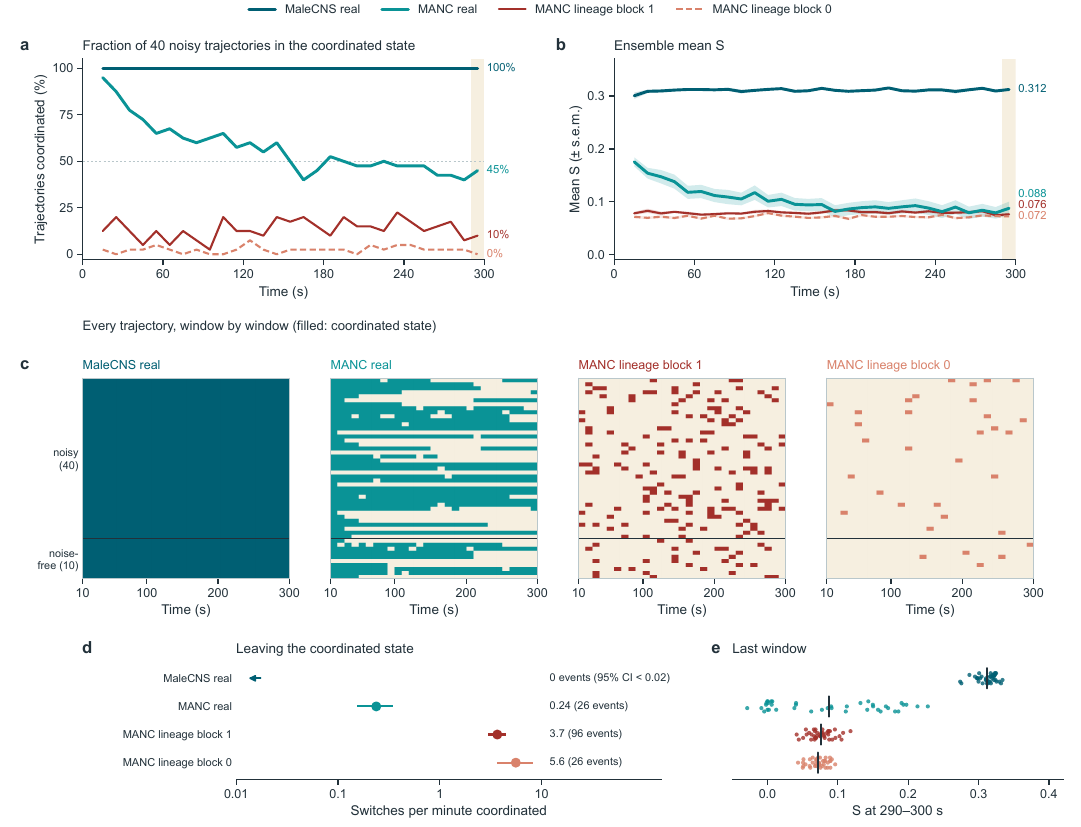}
\caption{\textbf{\capTitleEDB} \capBodyEDB}
\label{edfig:persist}
\end{figure}

\begin{figure}[p]
\figpage{183mm}{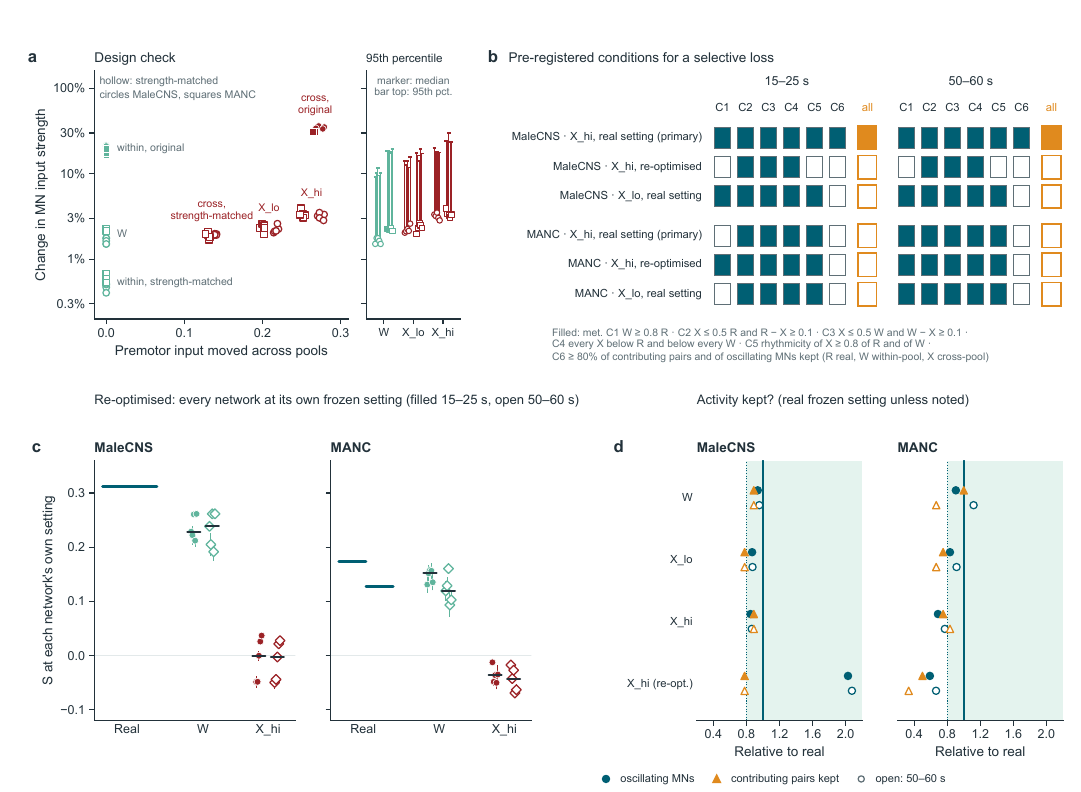}
\caption{\textbf{\capTitleEDC} \capBodyEDC}
\label{edfig:dose}
\end{figure}

\begin{figure}[p]
\figpage{183mm}{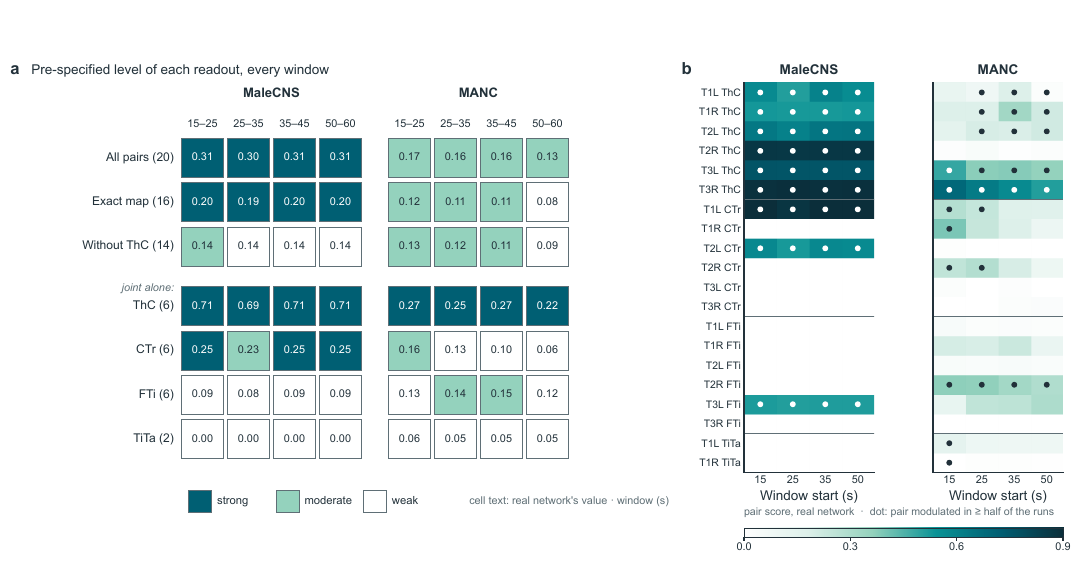}
\caption{\textbf{\capTitleEDD} \capBodyEDD}
\label{edfig:levels}
\end{figure}

\clearpage

\begin{table}[h]\centering\footnotesize
\caption{\textbf{Pre-specified claim levels in the 60-s confirmation runs.} Network values are means of 20 stochastic runs at each network's frozen setting; family values are medians over the five networks of a family. Strongest rewired network and smallest margin (real minus family median) are taken over the six main families. Strong: real above all 30 networks and every margin $\geq0.15$; moderate: real above every network in at least five families and above every family median.}
\label{tab:levels}
\begin{tabular}{llccccl}\toprule
Connectome & Window (s) & Real $S$ & \shortstack{Strongest rewired\\(family)} & \shortstack{Smallest margin\\(family)} & \shortstack{Families\\all below} & Level \\\midrule
MaleCNS & 15--25 & $0.312$ & $0.144$ (Lineage block) & $0.271$ (Lineage block) & 6/6 & strong \\
MaleCNS & 25--35 & $0.302$ & $0.146$ (Lineage block) & $0.260$ (Lineage block) & 6/6 & strong \\
MaleCNS & 35--45 & $0.313$ & $0.144$ (Lineage block) & $0.274$ (Lineage block) & 6/6 & strong \\
MaleCNS & 50--60 & $0.312$ & $0.147$ (Lineage block) & $0.277$ (Lineage block) & 6/6 & strong \\
\addlinespace
MANC & 15--25 & $0.173$ & $0.084$ (Lineage block) & $0.134$ (Lineage block) & 6/6 & moderate \\
MANC & 25--35 & $0.161$ & $0.080$ (Lineage block) & $0.114$ (Lineage block) & 6/6 & moderate \\
MANC & 35--45 & $0.160$ & $0.078$ (Lineage block) & $0.117$ (Lineage block) & 6/6 & moderate \\
MANC & 50--60 & $0.128$ & $0.080$ (Lineage block) & $0.086$ (Lineage block) & 6/6 & moderate \\
\bottomrule\end{tabular}\end{table}

\begin{table}[p]\centering\scriptsize
\caption{\textbf{Pre-specified conditions for a selective loss after motor-input reassignment.} $R$, real network; $W$ and $X$, medians over the five within-pool and five cross-pool networks (each a mean of 20 runs). Parameter rule: real, every network at the real network's frozen setting; own, every network at its own selected setting. Conditions: (1) $W\geq0.8R$; (2) $X\leq0.5R$ and $R-X\geq0.10$; (3) $X\leq0.5W$ and $W-X\geq0.10$; (4) every cross-pool network below the real network and below every within-pool network; (5) cross-pool rhythmicity at least 80\% of both references; (6) at least 80\% retention of modulated antagonistic pairs and of oscillating motor neurons, here assessed from the number of modulated pairs (the 60-s runs did not store pair identities; see Supplementary Table~\ref{tab:c6} for the pair-identity assessment).}
\label{tab:surgery}
\setlength{\tabcolsep}{3.2pt}
\begin{tabular}{llllccccccccccc}\toprule
Connectome & Reassignment & Rule & Window (s) & $R$ & $W$ & $X$ & $X/R$ & (1) & (2) & (3) & (4) & (5) & (6) & All six \\\midrule
MaleCNS & Original & real & 15--25 & $0.312$ & $0.298$ & $-0.071$ & $-0.23$ & \ding{51} & \ding{51} & \ding{51} & \ding{51} & \ding{51} & \ding{51} & \ding{51} \\
MaleCNS & Original & real & 25--35 & $0.302$ & $0.294$ & $-0.071$ & $-0.24$ & \ding{51} & \ding{51} & \ding{51} & \ding{51} & \ding{51} & \ding{51} & \ding{51} \\
MaleCNS & Original & real & 35--45 & $0.313$ & $0.295$ & $-0.070$ & $-0.22$ & \ding{51} & \ding{51} & \ding{51} & \ding{51} & \ding{51} & \ding{51} & \ding{51} \\
MaleCNS & Original & real & 50--60 & $0.312$ & $0.297$ & $-0.070$ & $-0.23$ & \ding{51} & \ding{51} & \ding{51} & \ding{51} & \ding{51} & \ding{51} & \ding{51} \\
MaleCNS & Original & own & 15--25 & $0.312$ & $0.283$ & $-0.029$ & $-0.09$ & \ding{51} & \ding{51} & \ding{51} & \ding{51} & \ding{55} & \ding{51} & \ding{55} \\
MaleCNS & Original & own & 25--35 & $0.302$ & $0.294$ & $-0.028$ & $-0.09$ & \ding{51} & \ding{51} & \ding{51} & \ding{51} & \ding{55} & \ding{51} & \ding{55} \\
MaleCNS & Original & own & 35--45 & $0.313$ & $0.290$ & $-0.026$ & $-0.08$ & \ding{51} & \ding{51} & \ding{51} & \ding{51} & \ding{55} & \ding{51} & \ding{55} \\
MaleCNS & Original & own & 50--60 & $0.312$ & $0.287$ & $-0.018$ & $-0.06$ & \ding{51} & \ding{51} & \ding{51} & \ding{51} & \ding{55} & \ding{51} & \ding{55} \\
\addlinespace
MaleCNS & Strength-matched & real & 15--25 & $0.312$ & $0.297$ & $0.170$ & $0.55$ & \ding{51} & \ding{55} & \ding{55} & \ding{51} & \ding{51} & \ding{51} & \ding{55} \\
MaleCNS & Strength-matched & real & 25--35 & $0.302$ & $0.295$ & $0.163$ & $0.54$ & \ding{51} & \ding{55} & \ding{55} & \ding{51} & \ding{51} & \ding{51} & \ding{55} \\
MaleCNS & Strength-matched & real & 35--45 & $0.313$ & $0.311$ & $0.169$ & $0.54$ & \ding{51} & \ding{55} & \ding{55} & \ding{51} & \ding{51} & \ding{51} & \ding{55} \\
MaleCNS & Strength-matched & real & 50--60 & $0.312$ & $0.309$ & $0.163$ & $0.52$ & \ding{51} & \ding{55} & \ding{55} & \ding{51} & \ding{51} & \ding{51} & \ding{55} \\
MaleCNS & Strength-matched & own & 15--25 & $0.312$ & $0.285$ & $0.215$ & $0.69$ & \ding{51} & \ding{55} & \ding{55} & \ding{51} & \ding{55} & \ding{51} & \ding{55} \\
MaleCNS & Strength-matched & own & 25--35 & $0.302$ & $0.286$ & $0.216$ & $0.72$ & \ding{51} & \ding{55} & \ding{55} & \ding{51} & \ding{55} & \ding{51} & \ding{55} \\
MaleCNS & Strength-matched & own & 35--45 & $0.313$ & $0.285$ & $0.205$ & $0.65$ & \ding{51} & \ding{55} & \ding{55} & \ding{51} & \ding{55} & \ding{51} & \ding{55} \\
MaleCNS & Strength-matched & own & 50--60 & $0.312$ & $0.287$ & $0.211$ & $0.68$ & \ding{51} & \ding{55} & \ding{55} & \ding{51} & \ding{55} & \ding{51} & \ding{55} \\
\addlinespace
MANC & Original & real & 15--25 & $0.173$ & $0.140$ & $-0.056$ & $-0.32$ & \ding{51} & \ding{51} & \ding{51} & \ding{51} & \ding{51} & \ding{51} & \ding{51} \\
MANC & Original & real & 25--35 & $0.161$ & $0.127$ & $-0.048$ & $-0.30$ & \ding{55} & \ding{51} & \ding{51} & \ding{51} & \ding{55} & \ding{51} & \ding{55} \\
MANC & Original & real & 35--45 & $0.160$ & $0.137$ & $-0.069$ & $-0.43$ & \ding{51} & \ding{51} & \ding{51} & \ding{51} & \ding{55} & \ding{51} & \ding{55} \\
MANC & Original & real & 50--60 & $0.128$ & $0.108$ & $-0.051$ & $-0.40$ & \ding{51} & \ding{51} & \ding{51} & \ding{51} & \ding{51} & \ding{51} & \ding{51} \\
MANC & Original & own & 15--25 & $0.173$ & $0.150$ & $-0.029$ & $-0.17$ & \ding{51} & \ding{51} & \ding{51} & \ding{51} & \ding{51} & \ding{51} & \ding{51} \\
MANC & Original & own & 25--35 & $0.161$ & $0.137$ & $-0.029$ & $-0.18$ & \ding{51} & \ding{51} & \ding{51} & \ding{51} & \ding{51} & \ding{51} & \ding{51} \\
MANC & Original & own & 35--45 & $0.160$ & $0.127$ & $-0.027$ & $-0.17$ & \ding{55} & \ding{51} & \ding{51} & \ding{51} & \ding{51} & \ding{51} & \ding{55} \\
MANC & Original & own & 50--60 & $0.128$ & $0.112$ & $-0.036$ & $-0.28$ & \ding{51} & \ding{51} & \ding{51} & \ding{51} & \ding{51} & \ding{51} & \ding{51} \\
\addlinespace
MANC & Strength-matched & real & 15--25 & $0.173$ & $0.166$ & $0.094$ & $0.54$ & \ding{51} & \ding{55} & \ding{55} & \ding{51} & \ding{51} & \ding{51} & \ding{55} \\
MANC & Strength-matched & real & 25--35 & $0.161$ & $0.155$ & $0.083$ & $0.51$ & \ding{51} & \ding{55} & \ding{55} & \ding{51} & \ding{51} & \ding{51} & \ding{55} \\
MANC & Strength-matched & real & 35--45 & $0.160$ & $0.161$ & $0.081$ & $0.50$ & \ding{51} & \ding{55} & \ding{55} & \ding{51} & \ding{51} & \ding{51} & \ding{55} \\
MANC & Strength-matched & real & 50--60 & $0.128$ & $0.139$ & $0.065$ & $0.51$ & \ding{51} & \ding{55} & \ding{55} & \ding{51} & \ding{51} & \ding{51} & \ding{55} \\
MANC & Strength-matched & own & 15--25 & $0.173$ & $0.162$ & $0.107$ & $0.62$ & \ding{51} & \ding{55} & \ding{55} & \ding{55} & \ding{51} & \ding{51} & \ding{55} \\
MANC & Strength-matched & own & 25--35 & $0.161$ & $0.159$ & $0.136$ & $0.84$ & \ding{51} & \ding{55} & \ding{55} & \ding{55} & \ding{51} & \ding{51} & \ding{55} \\
MANC & Strength-matched & own & 35--45 & $0.160$ & $0.140$ & $0.101$ & $0.63$ & \ding{51} & \ding{55} & \ding{55} & \ding{55} & \ding{51} & \ding{51} & \ding{55} \\
MANC & Strength-matched & own & 50--60 & $0.128$ & $0.135$ & $0.077$ & $0.60$ & \ding{51} & \ding{55} & \ding{55} & \ding{55} & \ding{51} & \ding{51} & \ding{55} \\
\bottomrule\end{tabular}\end{table}

\begin{table}[p]\centering\scriptsize
\caption{\textbf{Pair-level readouts of the confirmation runs (second-round pre-registration).} Same trajectories as the 60-s confirmation (reproduced to within rounding error), every network at its own frozen setting; pairs failing the amplitude gate count as zero. Readouts without a dagger were pre-specified decision readouts; $^{\dagger}$, descriptive. Level: the rule of Supplementary Table~\ref{tab:levels} applied to each readout.}
\label{tab:readouts}
\renewcommand{\arraystretch}{0.86}
\begin{tabular}{lllcccll}\toprule
Connectome & Readout & Window (s) & Real & \shortstack{Strongest rewired\\(family)} & \shortstack{Smallest\\margin} & \shortstack{Families\\all below} & Level \\\midrule
MaleCNS & All pairs (20)$^{\dagger}$ & 15--25 & $0.312$ & $0.144$ (Lineage block) & $0.271$ & 6/6 & strong \\
MaleCNS & All pairs (20)$^{\dagger}$ & 25--35 & $0.302$ & $0.146$ (Lineage block) & $0.260$ & 6/6 & strong \\
MaleCNS & All pairs (20)$^{\dagger}$ & 35--45 & $0.313$ & $0.144$ (Lineage block) & $0.274$ & 6/6 & strong \\
MaleCNS & All pairs (20)$^{\dagger}$ & 50--60 & $0.312$ & $0.147$ (Lineage block) & $0.277$ & 6/6 & strong \\
\addlinespace[2pt]
MaleCNS & Exact map (16) & 15--25 & $0.200$ & $0.126$ (Degree) & $0.178$ & 6/6 & strong \\
MaleCNS & Exact map (16) & 25--35 & $0.190$ & $0.126$ (Degree) & $0.174$ & 6/6 & strong \\
MaleCNS & Exact map (16) & 35--45 & $0.201$ & $0.125$ (Degree) & $0.174$ & 6/6 & strong \\
MaleCNS & Exact map (16) & 50--60 & $0.200$ & $0.125$ (Degree) & $0.180$ & 6/6 & strong \\
\addlinespace[2pt]
MaleCNS & No ThC (14) & 15--25 & $0.143$ & $0.169$ (Degree) & $0.105$ & 5/6 & moderate \\
MaleCNS & No ThC (14) & 25--35 & $0.136$ & $0.169$ (Degree) & $0.107$ & 4/6 & weak \\
MaleCNS & No ThC (14) & 35--45 & $0.143$ & $0.168$ (Degree) & $0.098$ & 4/6 & weak \\
MaleCNS & No ThC (14) & 50--60 & $0.143$ & $0.169$ (Degree) & $0.107$ & 4/6 & weak \\
\addlinespace[2pt]
MaleCNS & ThC (6)$^{\dagger}$ & 15--25 & $0.705$ & $0.215$ (Lineage block) & $0.586$ & 6/6 & strong \\
MaleCNS & ThC (6)$^{\dagger}$ & 25--35 & $0.689$ & $0.215$ (Lineage block) & $0.560$ & 6/6 & strong \\
MaleCNS & ThC (6)$^{\dagger}$ & 35--45 & $0.710$ & $0.215$ (Lineage block) & $0.593$ & 6/6 & strong \\
MaleCNS & ThC (6)$^{\dagger}$ & 50--60 & $0.705$ & $0.219$ (Lineage block) & $0.587$ & 6/6 & strong \\
\addlinespace[2pt]
MaleCNS & CTr (6)$^{\dagger}$ & 15--25 & $0.249$ & $0.240$ (Degree) & $0.224$ & 6/6 & strong \\
MaleCNS & CTr (6)$^{\dagger}$ & 25--35 & $0.232$ & $0.240$ (Degree) & $0.208$ & 5/6 & moderate \\
MaleCNS & CTr (6)$^{\dagger}$ & 35--45 & $0.249$ & $0.239$ (Degree) & $0.228$ & 6/6 & strong \\
MaleCNS & CTr (6)$^{\dagger}$ & 50--60 & $0.248$ & $0.240$ (Degree) & $0.232$ & 6/6 & strong \\
\addlinespace[2pt]
MaleCNS & FTi (6)$^{\dagger}$ & 15--25 & $0.086$ & $0.273$ (Lineage block) & $0.043$ & 3/6 & weak \\
MaleCNS & FTi (6)$^{\dagger}$ & 25--35 & $0.085$ & $0.277$ (Lineage block) & $0.029$ & 3/6 & weak \\
MaleCNS & FTi (6)$^{\dagger}$ & 35--45 & $0.085$ & $0.275$ (Lineage block) & $0.024$ & 3/6 & weak \\
MaleCNS & FTi (6)$^{\dagger}$ & 50--60 & $0.085$ & $0.272$ (Lineage block) & $0.032$ & 3/6 & weak \\
\addlinespace[2pt]
MaleCNS & TiTa (2)$^{\dagger}$ & 15--25 & $0.000$ & $0.371$ (Degree) & $-0.113$ & 0/6 & weak \\
MaleCNS & TiTa (2)$^{\dagger}$ & 25--35 & $0.000$ & $0.373$ (Degree) & $-0.090$ & 0/6 & weak \\
MaleCNS & TiTa (2)$^{\dagger}$ & 35--45 & $0.000$ & $0.369$ (Degree) & $-0.102$ & 0/6 & weak \\
MaleCNS & TiTa (2)$^{\dagger}$ & 50--60 & $0.000$ & $0.374$ (Degree) & $-0.106$ & 0/6 & weak \\
\addlinespace[2pt]
MANC & All pairs (20)$^{\dagger}$ & 15--25 & $0.173$ & $0.084$ (Lineage block) & $0.134$ & 6/6 & moderate \\
MANC & All pairs (20)$^{\dagger}$ & 25--35 & $0.161$ & $0.080$ (Lineage block) & $0.114$ & 6/6 & moderate \\
MANC & All pairs (20)$^{\dagger}$ & 35--45 & $0.160$ & $0.078$ (Lineage block) & $0.117$ & 6/6 & moderate \\
MANC & All pairs (20)$^{\dagger}$ & 50--60 & $0.128$ & $0.080$ (Lineage block) & $0.086$ & 6/6 & moderate \\
\addlinespace[2pt]
MANC & Exact map (16) & 15--25 & $0.122$ & $0.134$ (Degree) & $0.105$ & 5/6 & moderate \\
MANC & Exact map (16) & 25--35 & $0.113$ & $0.136$ (Degree) & $0.112$ & 5/6 & moderate \\
MANC & Exact map (16) & 35--45 & $0.108$ & $0.124$ (Degree) & $0.088$ & 5/6 & moderate \\
MANC & Exact map (16) & 50--60 & $0.075$ & $0.127$ (Degree) & $0.067$ & 4/6 & weak \\
\addlinespace[2pt]
MANC & No ThC (14) & 15--25 & $0.131$ & $0.121$ (Degree) & $0.107$ & 6/6 & moderate \\
MANC & No ThC (14) & 25--35 & $0.123$ & $0.128$ (Degree) & $0.086$ & 5/6 & moderate \\
MANC & No ThC (14) & 35--45 & $0.112$ & $0.113$ (Degree) & $0.082$ & 5/6 & moderate \\
MANC & No ThC (14) & 50--60 & $0.088$ & $0.115$ (Degree) & $0.063$ & 4/6 & weak \\
\addlinespace[2pt]
MANC & ThC (6)$^{\dagger}$ & 15--25 & $0.271$ & $0.105$ (Lineage block) & $0.197$ & 6/6 & strong \\
MANC & ThC (6)$^{\dagger}$ & 25--35 & $0.249$ & $0.118$ (Lineage block) & $0.195$ & 6/6 & strong \\
MANC & ThC (6)$^{\dagger}$ & 35--45 & $0.275$ & $0.110$ (Lineage block) & $0.219$ & 6/6 & strong \\
MANC & ThC (6)$^{\dagger}$ & 50--60 & $0.222$ & $0.105$ (Lineage block) & $0.178$ & 6/6 & strong \\
\addlinespace[2pt]
MANC & CTr (6)$^{\dagger}$ & 15--25 & $0.159$ & $0.145$ (Lineage block) & $0.033$ & 6/6 & moderate \\
MANC & CTr (6)$^{\dagger}$ & 25--35 & $0.127$ & $0.158$ (Lineage block) & $-0.002$ & 4/6 & weak \\
MANC & CTr (6)$^{\dagger}$ & 35--45 & $0.097$ & $0.141$ (Lineage block) & $-0.030$ & 4/6 & weak \\
MANC & CTr (6)$^{\dagger}$ & 50--60 & $0.064$ & $0.152$ (Lineage block) & $-0.066$ & 4/6 & weak \\
\addlinespace[2pt]
MANC & FTi (6)$^{\dagger}$ & 15--25 & $0.128$ & $0.150$ (Degree) & $0.119$ & 4/6 & weak \\
MANC & FTi (6)$^{\dagger}$ & 25--35 & $0.143$ & $0.165$ (Lineage block) & $0.120$ & 5/6 & moderate \\
MANC & FTi (6)$^{\dagger}$ & 35--45 & $0.146$ & $0.160$ (Lineage block) & $0.123$ & 5/6 & moderate \\
MANC & FTi (6)$^{\dagger}$ & 50--60 & $0.125$ & $0.157$ (Lineage block) & $0.114$ & 4/6 & weak \\
\addlinespace[2pt]
MANC & TiTa (2)$^{\dagger}$ & 15--25 & $0.057$ & $0.452$ (Lineage block) & $0.005$ & 2/6 & weak \\
MANC & TiTa (2)$^{\dagger}$ & 25--35 & $0.053$ & $0.452$ (Lineage block) & $-0.008$ & 2/6 & weak \\
MANC & TiTa (2)$^{\dagger}$ & 35--45 & $0.050$ & $0.450$ (Lineage block) & $0.004$ & 2/6 & weak \\
MANC & TiTa (2)$^{\dagger}$ & 50--60 & $0.049$ & $0.445$ (Lineage block) & $0.000$ & 2/6 & weak \\
\bottomrule\end{tabular}\end{table}

\begin{table}[p]\centering\footnotesize
\caption{\textbf{Condition 6 assessed by pair identity (reassigned networks at the real frozen setting).} Contributing pairs: pairs that pass the amplitude gate in at least half of the real network's 20 runs. Pair retention: fraction of these pairs that still pass the gate in at least half of a reassigned network's runs (median over five networks). Oscillation retention: fraction of oscillating motor neurons relative to the real network (median over five networks). Condition met: both medians $\geq0.8$.}
\label{tab:c6}
\begin{tabular}{lllcccc}\toprule
Connectome & Reassignment & Window (s) & \shortstack{Contributing\\pairs} & \shortstack{Pair\\retention} & \shortstack{Oscillation\\retention} & Met \\\midrule
MaleCNS & Original cross-pool & 15--25 & 9 & 0.89 & 0.97 & \ding{51} \\
MaleCNS & Original cross-pool & 25--35 & 9 & 0.89 & 1.04 & \ding{51} \\
MaleCNS & Original cross-pool & 35--45 & 9 & 1.00 & 1.00 & \ding{51} \\
MaleCNS & Original cross-pool & 50--60 & 9 & 0.89 & 1.02 & \ding{51} \\
\addlinespace
MaleCNS & Original within-pool & 15--25 & 9 & 1.00 & 0.99 & \ding{51} \\
MaleCNS & Original within-pool & 25--35 & 9 & 0.89 & 1.05 & \ding{51} \\
MaleCNS & Original within-pool & 35--45 & 9 & 1.00 & 0.99 & \ding{51} \\
MaleCNS & Original within-pool & 50--60 & 9 & 1.00 & 1.03 & \ding{51} \\
\addlinespace
MaleCNS & Strength-matched cross-pool & 15--25 & 9 & 0.89 & 0.87 & \ding{51} \\
MaleCNS & Strength-matched cross-pool & 25--35 & 9 & 0.89 & 0.93 & \ding{51} \\
MaleCNS & Strength-matched cross-pool & 35--45 & 9 & 0.89 & 0.88 & \ding{51} \\
MaleCNS & Strength-matched cross-pool & 50--60 & 9 & 0.89 & 0.90 & \ding{51} \\
\addlinespace
MaleCNS & Strength-matched within-pool & 15--25 & 9 & 1.00 & 0.96 & \ding{51} \\
MaleCNS & Strength-matched within-pool & 25--35 & 9 & 1.00 & 1.00 & \ding{51} \\
MaleCNS & Strength-matched within-pool & 35--45 & 9 & 1.00 & 0.97 & \ding{51} \\
MaleCNS & Strength-matched within-pool & 50--60 & 9 & 1.00 & 0.98 & \ding{51} \\
\addlinespace
MANC & Original cross-pool & 15--25 & 8 & 0.88 & 0.96 & \ding{51} \\
MANC & Original cross-pool & 25--35 & 8 & 0.75 & 1.00 & \ding{55} \\
MANC & Original cross-pool & 35--45 & 6 & 0.83 & 1.03 & \ding{51} \\
MANC & Original cross-pool & 50--60 & 6 & 0.67 & 1.01 & \ding{55} \\
\addlinespace
MANC & Original within-pool & 15--25 & 8 & 1.00 & 0.93 & \ding{51} \\
MANC & Original within-pool & 25--35 & 8 & 0.75 & 0.89 & \ding{55} \\
MANC & Original within-pool & 35--45 & 6 & 0.83 & 0.92 & \ding{51} \\
MANC & Original within-pool & 50--60 & 6 & 0.83 & 1.00 & \ding{51} \\
\addlinespace
MANC & Strength-matched cross-pool & 15--25 & 8 & 0.88 & 0.85 & \ding{51} \\
MANC & Strength-matched cross-pool & 25--35 & 8 & 0.62 & 0.80 & \ding{55} \\
MANC & Strength-matched cross-pool & 35--45 & 6 & 0.67 & 0.89 & \ding{55} \\
MANC & Strength-matched cross-pool & 50--60 & 6 & 0.67 & 0.95 & \ding{55} \\
\addlinespace
MANC & Strength-matched within-pool & 15--25 & 8 & 1.00 & 0.99 & \ding{51} \\
MANC & Strength-matched within-pool & 25--35 & 8 & 0.75 & 1.00 & \ding{55} \\
MANC & Strength-matched within-pool & 35--45 & 6 & 1.00 & 1.11 & \ding{51} \\
MANC & Strength-matched within-pool & 50--60 & 6 & 1.00 & 1.17 & \ding{51} \\
\bottomrule\end{tabular}\end{table}

\begin{table}[p]\centering\scriptsize
\caption{\textbf{Dose-matched reassignment: pre-specified conditions (second-round pre-registration).} Cross-pool arms X$_{\mathrm{hi}}$ (dose-matched) and X$_{\mathrm{lo}}$ (lower dose, matched to the within-pool control W in the fraction of new edges), five networks each, exchanges restricted to edges of the same sign and synapse-count decile. Rule: real, every network at the real network's frozen setting; own, every network at its own re-optimised setting; $^{\ast}$, primary test. $R$, real network; $W$ and $X$, medians of five network means (20 runs each). Rhythm ratio, cross-pool rhythmicity relative to the real network. Pairs kept and oscillation kept: condition 6 by pair identity (medians of five networks, relative to the real network). Conditions (1)--(6) as in Supplementary Table~\ref{tab:surgery}.}
\label{tab:exp2}
\setlength{\tabcolsep}{2.1pt}
\begin{tabular}{lllcccccccccccccc}\toprule
Connectome & Arm, rule & Window (s) & $R$ & $W$ & $X$ & $X/R$ & \shortstack{Rhythm\\ratio} & \shortstack{Pairs\\kept} & \shortstack{Osc.\\kept} & (1) & (2) & (3) & (4) & (5) & (6) & All six \\\midrule
MaleCNS & X$_{\mathrm{hi}}$, real$^{\ast}$ & 15--25 & $0.312$ & $0.277$ & $-0.076$ & $-0.24$ & $0.97$ & 0.89 & 0.85 & \ding{51} & \ding{51} & \ding{51} & \ding{51} & \ding{51} & \ding{51} & \ding{51} \\
MaleCNS & X$_{\mathrm{hi}}$, real$^{\ast}$ & 25--35 & $0.302$ & $0.272$ & $-0.075$ & $-0.25$ & $1.09$ & 0.89 & 0.87 & \ding{51} & \ding{51} & \ding{51} & \ding{51} & \ding{51} & \ding{51} & \ding{51} \\
MaleCNS & X$_{\mathrm{hi}}$, real$^{\ast}$ & 35--45 & $0.313$ & $0.270$ & $-0.079$ & $-0.25$ & $0.78$ & 0.89 & 0.85 & \ding{51} & \ding{51} & \ding{51} & \ding{51} & \ding{55} & \ding{51} & \ding{55} \\
MaleCNS & X$_{\mathrm{hi}}$, real$^{\ast}$ & 50--60 & $0.312$ & $0.275$ & $-0.082$ & $-0.26$ & $0.92$ & 0.89 & 0.86 & \ding{51} & \ding{51} & \ding{51} & \ding{51} & \ding{51} & \ding{51} & \ding{51} \\
\addlinespace
MaleCNS & X$_{\mathrm{hi}}$, own & 15--25 & $0.312$ & $0.229$ & $-0.000$ & $-0.00$ & $0.42$ & 0.78 & 2.03 & \ding{55} & \ding{51} & \ding{51} & \ding{51} & \ding{55} & \ding{55} & \ding{55} \\
MaleCNS & X$_{\mathrm{hi}}$, own & 25--35 & $0.302$ & $0.239$ & $0.009$ & $0.03$ & $0.44$ & 0.78 & 2.10 & \ding{55} & \ding{51} & \ding{51} & \ding{51} & \ding{55} & \ding{55} & \ding{55} \\
MaleCNS & X$_{\mathrm{hi}}$, own & 35--45 & $0.313$ & $0.227$ & $0.014$ & $0.04$ & $0.37$ & 0.78 & 1.96 & \ding{55} & \ding{51} & \ding{51} & \ding{51} & \ding{55} & \ding{55} & \ding{55} \\
MaleCNS & X$_{\mathrm{hi}}$, own & 50--60 & $0.312$ & $0.239$ & $-0.003$ & $-0.01$ & $0.38$ & 0.78 & 2.08 & \ding{55} & \ding{51} & \ding{51} & \ding{51} & \ding{55} & \ding{55} & \ding{55} \\
\addlinespace
MaleCNS & X$_{\mathrm{lo}}$, real & 15--25 & $0.312$ & $0.277$ & $0.039$ & $0.13$ & $0.96$ & 0.78 & 0.87 & \ding{51} & \ding{51} & \ding{51} & \ding{51} & \ding{51} & \ding{55} & \ding{55} \\
MaleCNS & X$_{\mathrm{lo}}$, real & 25--35 & $0.302$ & $0.272$ & $0.038$ & $0.13$ & $1.08$ & 0.78 & 0.89 & \ding{51} & \ding{51} & \ding{51} & \ding{51} & \ding{51} & \ding{55} & \ding{55} \\
MaleCNS & X$_{\mathrm{lo}}$, real & 35--45 & $0.313$ & $0.270$ & $0.041$ & $0.13$ & $0.87$ & 0.78 & 0.86 & \ding{51} & \ding{51} & \ding{51} & \ding{51} & \ding{51} & \ding{55} & \ding{55} \\
MaleCNS & X$_{\mathrm{lo}}$, real & 50--60 & $0.312$ & $0.275$ & $0.041$ & $0.13$ & $1.05$ & 0.78 & 0.87 & \ding{51} & \ding{51} & \ding{51} & \ding{51} & \ding{51} & \ding{55} & \ding{55} \\
\addlinespace
MANC & X$_{\mathrm{hi}}$, real$^{\ast}$ & 15--25 & $0.173$ & $0.135$ & $-0.096$ & $-0.56$ & $1.58$ & 0.75 & 0.69 & \ding{55} & \ding{51} & \ding{51} & \ding{51} & \ding{51} & \ding{55} & \ding{55} \\
MANC & X$_{\mathrm{hi}}$, real$^{\ast}$ & 25--35 & $0.161$ & $0.128$ & $-0.092$ & $-0.57$ & $1.47$ & 0.62 & 0.69 & \ding{55} & \ding{51} & \ding{51} & \ding{51} & \ding{51} & \ding{55} & \ding{55} \\
MANC & X$_{\mathrm{hi}}$, real$^{\ast}$ & 35--45 & $0.160$ & $0.136$ & $-0.087$ & $-0.54$ & $1.50$ & 0.67 & 0.75 & \ding{51} & \ding{51} & \ding{51} & \ding{51} & \ding{51} & \ding{55} & \ding{55} \\
MANC & X$_{\mathrm{hi}}$, real$^{\ast}$ & 50--60 & $0.128$ & $0.119$ & $-0.088$ & $-0.69$ & $1.36$ & 0.83 & 0.77 & \ding{51} & \ding{51} & \ding{51} & \ding{51} & \ding{51} & \ding{55} & \ding{55} \\
\addlinespace
MANC & X$_{\mathrm{hi}}$, own & 15--25 & $0.173$ & $0.152$ & $-0.036$ & $-0.21$ & $1.00$ & 0.50 & 0.59 & \ding{51} & \ding{51} & \ding{51} & \ding{51} & \ding{51} & \ding{55} & \ding{55} \\
MANC & X$_{\mathrm{hi}}$, own & 25--35 & $0.161$ & $0.128$ & $-0.041$ & $-0.25$ & $0.80$ & 0.50 & 0.59 & \ding{55} & \ding{51} & \ding{51} & \ding{51} & \ding{55} & \ding{55} & \ding{55} \\
MANC & X$_{\mathrm{hi}}$, own & 35--45 & $0.160$ & $0.132$ & $-0.042$ & $-0.26$ & $1.02$ & 0.50 & 0.58 & \ding{51} & \ding{51} & \ding{51} & \ding{51} & \ding{51} & \ding{55} & \ding{55} \\
MANC & X$_{\mathrm{hi}}$, own & 50--60 & $0.128$ & $0.119$ & $-0.043$ & $-0.33$ & $0.97$ & 0.33 & 0.66 & \ding{51} & \ding{51} & \ding{51} & \ding{51} & \ding{51} & \ding{55} & \ding{55} \\
\addlinespace
MANC & X$_{\mathrm{lo}}$, real & 15--25 & $0.173$ & $0.135$ & $0.024$ & $0.14$ & $1.21$ & 0.75 & 0.83 & \ding{55} & \ding{51} & \ding{51} & \ding{51} & \ding{51} & \ding{55} & \ding{55} \\
MANC & X$_{\mathrm{lo}}$, real & 25--35 & $0.161$ & $0.128$ & $0.011$ & $0.07$ & $1.11$ & 0.75 & 0.83 & \ding{55} & \ding{51} & \ding{51} & \ding{51} & \ding{51} & \ding{55} & \ding{55} \\
MANC & X$_{\mathrm{lo}}$, real & 35--45 & $0.160$ & $0.136$ & $0.004$ & $0.02$ & $1.08$ & 0.67 & 0.86 & \ding{51} & \ding{51} & \ding{51} & \ding{51} & \ding{51} & \ding{55} & \ding{55} \\
MANC & X$_{\mathrm{lo}}$, real & 50--60 & $0.128$ & $0.119$ & $-0.001$ & $-0.01$ & $1.06$ & 0.67 & 0.91 & \ding{51} & \ding{51} & \ding{51} & \ding{51} & \ding{51} & \ding{55} & \ding{55} \\
\bottomrule\end{tabular}\end{table}

\begin{table}[h]\centering\footnotesize
\caption{\textbf{300-s simulations (second-round pre-registration).} Noisy trajectories (20 continuing the confirmation runs, 20 new) at each network's frozen setting. Coordinated: a 10-s window with $S\geq0.10$ and at least six modulated pairs. Switch: a coordinated window followed by at least three non-coordinated windows. Rate: switches per minute spent coordinated, exact Poisson 95\% interval. Post hoc (not pre-registered): the difference in mean $S$ at 290--300~s between the real MANC network and lineage-block network 1 was 0.011 (bootstrap 95\% interval -0.013 to 0.036).}
\label{tab:exp3}
\begin{tabular}{lcccccc}\toprule
Network & Runs & \shortstack{Mean $S$\\290--300 s} & \shortstack{Coordinated\\at 290--300 s} & Switches & \shortstack{Minutes\\coordinated} & \shortstack{Switches per minute\\coordinated (95\% CI)} \\\midrule
MaleCNS real & 40 & $0.312$ & 100\% & 0 & 193.3 & 0.00 (0.00--0.02) \\
MANC real & 40 & $0.088$ & 45\% & 26 & 109.5 & 0.24 (0.16--0.35) \\
MANC lineage block 1 & 40 & $0.076$ & 10\% & 96 & 26.2 & 3.67 (2.97--4.48) \\
MANC lineage block 0 & 40 & $0.072$ & 0\% & 26 & 4.7 & 5.57 (3.64--8.16) \\
\bottomrule\end{tabular}\end{table}

\begin{table}[p]\centering\footnotesize
\caption{\textbf{Thorax--coxa pairs with all assigned motor neurons or with exactly assigned motor neurons only (post hoc).} Pair score of the real network (mean of 20 runs) and of the 30 main rewired networks (each at its own frozen setting). In both connectomes the promotor-side pool of the front legs contains pleural promotor (exact) and sternal anterior rotator (approximate) motor neurons, and that of the middle and hind legs only sternal anterior rotator motor neurons; exact-only pairs therefore exist only in the front legs.}
\label{tab:thc}
\begin{tabular}{llllccc}\toprule
Connectome & Pools & Window (s) & Pair & Real & \shortstack{Rewired\\maximum} & \shortstack{Rewired\\$\geq$ real} \\\midrule
MaleCNS & All assigned & 15--25 & T1L ThC & $0.567$ & $0.662$ & 2/30 \\
MaleCNS & All assigned & 15--25 & T1R ThC & $0.527$ & $0.494$ & 0/30 \\
MaleCNS & All assigned & 15--25 & T2L ThC & $0.630$ & $0.413$ & 0/30 \\
MaleCNS & All assigned & 15--25 & T2R ThC & $0.865$ & $0.874$ & 1/30 \\
MaleCNS & All assigned & 15--25 & T3L ThC & $0.759$ & $0.664$ & 0/30 \\
MaleCNS & All assigned & 15--25 & T3R ThC & $0.883$ & $0.597$ & 0/30 \\
MaleCNS & All assigned & 50--60 & T1L ThC & $0.561$ & $0.642$ & 2/30 \\
MaleCNS & All assigned & 50--60 & T1R ThC & $0.525$ & $0.482$ & 0/30 \\
MaleCNS & All assigned & 50--60 & T2L ThC & $0.645$ & $0.417$ & 0/30 \\
MaleCNS & All assigned & 50--60 & T2R ThC & $0.863$ & $0.873$ & 1/30 \\
MaleCNS & All assigned & 50--60 & T3L ThC & $0.753$ & $0.646$ & 0/30 \\
MaleCNS & All assigned & 50--60 & T3R ThC & $0.886$ & $0.581$ & 0/30 \\
\addlinespace
MaleCNS & Exact only & 15--25 & T1L ThC & $0.718$ & $0.458$ & 0/30 \\
MaleCNS & Exact only & 15--25 & T1R ThC & $0.477$ & $0.695$ & 1/30 \\
MaleCNS & Exact only & 50--60 & T1L ThC & $0.717$ & $0.467$ & 0/30 \\
MaleCNS & Exact only & 50--60 & T1R ThC & $0.479$ & $0.700$ & 1/30 \\
\addlinespace
MANC & All assigned & 15--25 & T1L ThC & $0.117$ & $0.613$ & 5/30 \\
MANC & All assigned & 15--25 & T1R ThC & $0.179$ & $0.375$ & 6/30 \\
MANC & All assigned & 15--25 & T2L ThC & $0.143$ & $0.162$ & 1/30 \\
MANC & All assigned & 15--25 & T2R ThC & $0.015$ & $0.315$ & 10/30 \\
MANC & All assigned & 15--25 & T3L ThC & $0.484$ & $0.479$ & 0/30 \\
MANC & All assigned & 15--25 & T3R ThC & $0.688$ & $0.293$ & 0/30 \\
MANC & All assigned & 50--60 & T1L ThC & $0.020$ & $0.617$ & 15/30 \\
MANC & All assigned & 50--60 & T1R ThC & $0.212$ & $0.407$ & 4/30 \\
MANC & All assigned & 50--60 & T2L ThC & $0.212$ & $0.081$ & 0/30 \\
MANC & All assigned & 50--60 & T2R ThC & $0.018$ & $0.325$ & 8/30 \\
MANC & All assigned & 50--60 & T3L ThC & $0.360$ & $0.480$ & 1/30 \\
MANC & All assigned & 50--60 & T3R ThC & $0.508$ & $0.294$ & 0/30 \\
\addlinespace
MANC & Exact only & 15--25 & T1L ThC & $0.032$ & $0.325$ & 4/30 \\
MANC & Exact only & 15--25 & T1R ThC & $0.147$ & $0.806$ & 6/30 \\
MANC & Exact only & 50--60 & T1L ThC & $-0.002$ & $0.280$ & 17/30 \\
MANC & Exact only & 50--60 & T1R ThC & $0.047$ & $0.797$ & 10/30 \\
\bottomrule\end{tabular}\end{table}

\begin{table}[h]\centering\footnotesize
\caption{\textbf{Antagonistic pairs that are active are antiphasic in the real networks (post hoc).} Active pair: both pools pass the amplitude gate in at least half of a network's 20 runs (the contributing pairs of condition 6). Antiphasic: mean extensor--flexor correlation over the gated runs below zero. Rewired: pairs pooled over the 30 main rewired networks, each at its own frozen setting. Mean correlation of each active pair of the real networks at 15--25~s: MaleCNS: T1L CTr $-0.92$; T2L CTr $-0.57$; T3L FTi $-0.51$; T1L ThC $-0.57$; T1R ThC $-0.53$; T2L ThC $-0.66$; T2R ThC $-0.86$; T3L ThC $-0.76$; T3R ThC $-0.88$; MANC: T1L CTr $-0.35$; T1R CTr $-0.49$; T2R CTr $-0.46$; T2R FTi $-0.39$; T3L ThC $-0.51$; T3R ThC $-0.72$; T1L TiTa $-0.21$; T1R TiTa $0.04$.}
\label{tab:r1}
\begin{tabular}{llccccc}\toprule
 & & \multicolumn{2}{c}{Real network} & \multicolumn{3}{c}{Rewired networks (30)} \\\cmidrule(lr){3-4}\cmidrule(lr){5-7}
Connectome & Window (s) & Active pairs & Antiphasic & Active pairs & Antiphasic & Fraction \\\midrule
MaleCNS & 15--25 & 9 & 9 & 428 & 207 & 48\% \\
MaleCNS & 50--60 & 9 & 9 & 425 & 219 & 52\% \\
MANC & 15--25 & 8 & 7 & 458 & 232 & 51\% \\
MANC & 50--60 & 6 & 6 & 455 & 219 & 48\% \\
\bottomrule\end{tabular}\end{table}

\begin{table}[p]\centering\footnotesize
\caption{\textbf{Opposite-signed influence behind the reciprocal-innervation index (post hoc).} For each antagonistic pair of the real networks: $f^{\mathrm{both}}$, the share of the influence on the two pools carried by neurons that influence both (Eq.~(\ref{eq:fboth})); $f^{\mathrm{opp}}$, the opposite-signed share of their influence (Eq.~(\ref{eq:fopp})); and the largest $f^{\mathrm{opp}}$ among the 30 main rewired networks. $^{\ast}$, real value above all 30 rewired networks. Influence as in the reciprocal-innervation index (one- plus two-step, signed synapse counts). Network-level values (medians over pairs): MaleCNS, real network 0.96 and 0.88, rewired networks $f^{\mathrm{opp}}$ 0.16--0.49 (30 networks); $f^{\mathrm{opp}}>0.5$ at 19 of 20 pairs and above all rewired networks at 18 of 20; MANC, real network 0.97 and 0.83, rewired networks $f^{\mathrm{opp}}$ 0.16--0.49 (30 networks); $f^{\mathrm{opp}}>0.5$ at 19 of 20 pairs and above all rewired networks at 18 of 20.}
\label{tab:fopp}
\begin{tabular}{lcccccc}\toprule
 & \multicolumn{3}{c}{MaleCNS} & \multicolumn{3}{c}{MANC} \\\cmidrule(lr){2-4}\cmidrule(lr){5-7}
Pair & $f^{\mathrm{both}}$ & $f^{\mathrm{opp}}$ & \shortstack{Rewired\\max. $f^{\mathrm{opp}}$} & $f^{\mathrm{both}}$ & $f^{\mathrm{opp}}$ & \shortstack{Rewired\\max. $f^{\mathrm{opp}}$} \\\midrule
T1L ThC & $0.97$ & $0.85$$^{\ast}$ & $0.52$ & $0.96$ & $0.87$$^{\ast}$ & $0.53$ \\
T1R ThC & $0.96$ & $0.88$$^{\ast}$ & $0.53$ & $0.96$ & $0.88$$^{\ast}$ & $0.55$ \\
T2L ThC & $0.98$ & $0.90$$^{\ast}$ & $0.59$ & $0.97$ & $0.83$$^{\ast}$ & $0.71$ \\
T2R ThC & $0.98$ & $0.87$$^{\ast}$ & $0.55$ & $0.96$ & $0.81$$^{\ast}$ & $0.78$ \\
T3L ThC & $0.98$ & $0.93$$^{\ast}$ & $0.51$ & $0.94$ & $0.91$$^{\ast}$ & $0.58$ \\
T3R ThC & $0.98$ & $0.92$$^{\ast}$ & $0.56$ & $0.98$ & $0.88$$^{\ast}$ & $0.56$ \\
T1L CTr & $0.92$ & $0.92$$^{\ast}$ & $0.51$ & $0.93$ & $0.90$$^{\ast}$ & $0.57$ \\
T1R CTr & $0.83$ & $0.92$$^{\ast}$ & $0.54$ & $0.95$ & $0.85$$^{\ast}$ & $0.57$ \\
T2L CTr & $0.97$ & $0.95$$^{\ast}$ & $0.62$ & $0.94$ & $0.91$$^{\ast}$ & $0.64$ \\
T2R CTr & $0.96$ & $0.95$$^{\ast}$ & $0.51$ & $0.97$ & $0.95$$^{\ast}$ & $0.53$ \\
T3L CTr & $0.83$ & $0.90$$^{\ast}$ & $0.56$ & $0.80$ & $0.82$$^{\ast}$ & $0.55$ \\
T3R CTr & $0.97$ & $0.94$$^{\ast}$ & $0.54$ & $0.98$ & $0.92$$^{\ast}$ & $0.53$ \\
T1L FTi & $0.98$ & $0.78$$^{\ast}$ & $0.55$ & $0.97$ & $0.76$$^{\ast}$ & $0.54$ \\
T1R FTi & $0.95$ & $0.72$$^{\ast}$ & $0.56$ & $0.98$ & $0.78$$^{\ast}$ & $0.63$ \\
T2L FTi & $0.97$ & $0.86$$^{\ast}$ & $0.54$ & $0.96$ & $0.77$$^{\ast}$ & $0.56$ \\
T2R FTi & $0.96$ & $0.78$$^{\ast}$ & $0.51$ & $0.98$ & $0.77$$^{\ast}$ & $0.63$ \\
T3L FTi & $0.96$ & $0.88$$^{\ast}$ & $0.62$ & $0.97$ & $0.77$$^{\ast}$ & $0.51$ \\
T3R FTi & $0.97$ & $0.87$$^{\ast}$ & $0.54$ & $0.99$ & $0.80$$^{\ast}$ & $0.69$ \\
T1L TiTa & $0.92$ & $0.57$\phantom{$^{\ast}$} & $0.61$ & $0.95$ & $0.30$\phantom{$^{\ast}$} & $0.55$ \\
T1R TiTa & $0.64$ & $0.45$\phantom{$^{\ast}$} & $0.65$ & $0.97$ & $0.52$\phantom{$^{\ast}$} & $0.64$ \\
\bottomrule\end{tabular}\end{table}

\end{document}